\documentclass{article}

\usepackage{microtype}
\usepackage{graphicx}
\usepackage{subcaption}
\usepackage{booktabs} 
\usepackage{multirow}

\usepackage{tikz}
\usetikzlibrary{matrix, positioning, arrows}
\usepackage{pgfplots}
\usepgfplotslibrary{external}
\usepgfplotslibrary{groupplots}
\usepackage{enumitem}
\setlist[itemize]{align=parleft,left=0pt..1em}

\usepackage{hyperref}

\usepackage[accepted]{icml2023}

\usepackage{amsmath}
\usepackage{amssymb}
\usepackage{mathtools}
\usepackage{amsthm}
\usepackage{xspace}

\usepackage[capitalize,noabbrev]{cleveref}

\DeclarePairedDelimiterX{\inner}[2]{\langle}{\rangle}{#1, #2}
\DeclarePairedDelimiterX{\inners2}[2]{\langle}{\rangle_{L^2(S^2)}}{#1, #2}

\theoremstyle{plain}

\theoremstyle{definition}

\theoremstyle{remark}

\usepackage[textsize=tiny]{todonotes}

\usepackage{xcolor}
\usepackage{physics}
\renewcommand\vb[1]{#1}

\icmltitlerunning{Spherical Fourier Neural Operators}

\begin{document}

\twocolumn[
\icmltitle{Spherical Fourier Neural Operators: \\ Learning Stable Dynamics on the Sphere}




\begin{icmlauthorlist}
\icmlauthor{Boris Bonev}{nvidia}
\icmlauthor{Thorsten Kurth}{nvidia}
\icmlauthor{Christian Hundt}{nvidia}
\icmlauthor{Jaideep Pathak}{nvidia}
\icmlauthor{Maximilian Baust}{nvidia}
\icmlauthor{Karthik Kashinath}{nvidia}
\icmlauthor{Anima Anandkumar}{nvidia,caltech}
\end{icmlauthorlist}

\icmlaffiliation{nvidia}{NVIDIA Corp., Santa Clara, USA}
\icmlaffiliation{caltech}{Caltech, Pasadena, USA}

\icmlcorrespondingauthor{Boris Bonev}{bbonev@nvidia.com}
\icmlcorrespondingauthor{Thorsten Kurth}{tkurth@nvidia.com}
\icmlcorrespondingauthor{Christian Hundt}{chundt@nvidia.com}

\icmlkeywords{Spherical Fourier Neural Operators, Spherical Geometry, PDEs, Equivariance, Weather Forecasting, Shallow Water Equations, Machine Learning, ICML}

\vskip 0.3in
]



\printAffiliationsAndNotice{}  

\begin{abstract}

Fourier Neural Operators (FNOs) have proven to be an efficient and effective method for resolution-independent operator learning in a broad variety of application areas across scientific machine learning.
A key reason for their success is their ability to accurately model long-range dependencies in spatio-temporal data by learning global convolutions in a computationally efficient manner.
To this end, FNOs rely on the discrete Fourier transform (DFT), however, DFTs cause visual and spectral artifacts as well as pronounced dissipation when learning operators in spherical coordinates since they incorrectly assume a flat geometry.
To overcome this limitation, we generalize FNOs on the sphere, introducing Spherical FNOs (SFNOs) for learning operators on spherical geometries.
We apply SFNOs to forecasting atmospheric dynamics, and demonstrate stable auto\-regressive rollouts for a year of simulated time (1,460 steps), while retaining physically plausible dynamics. The SFNO has important implications for machine learning-based simulation of climate dynamics that could eventually help accelerate our response to climate change.

\end{abstract}

\section{Introduction}
\label{sec:introduction}

\begin{figure}[!htb]
    \centering
    \begin{subfigure}{.48\linewidth}
        \includegraphics[width=\linewidth, trim={14px 14px 14px 14px}, clip]{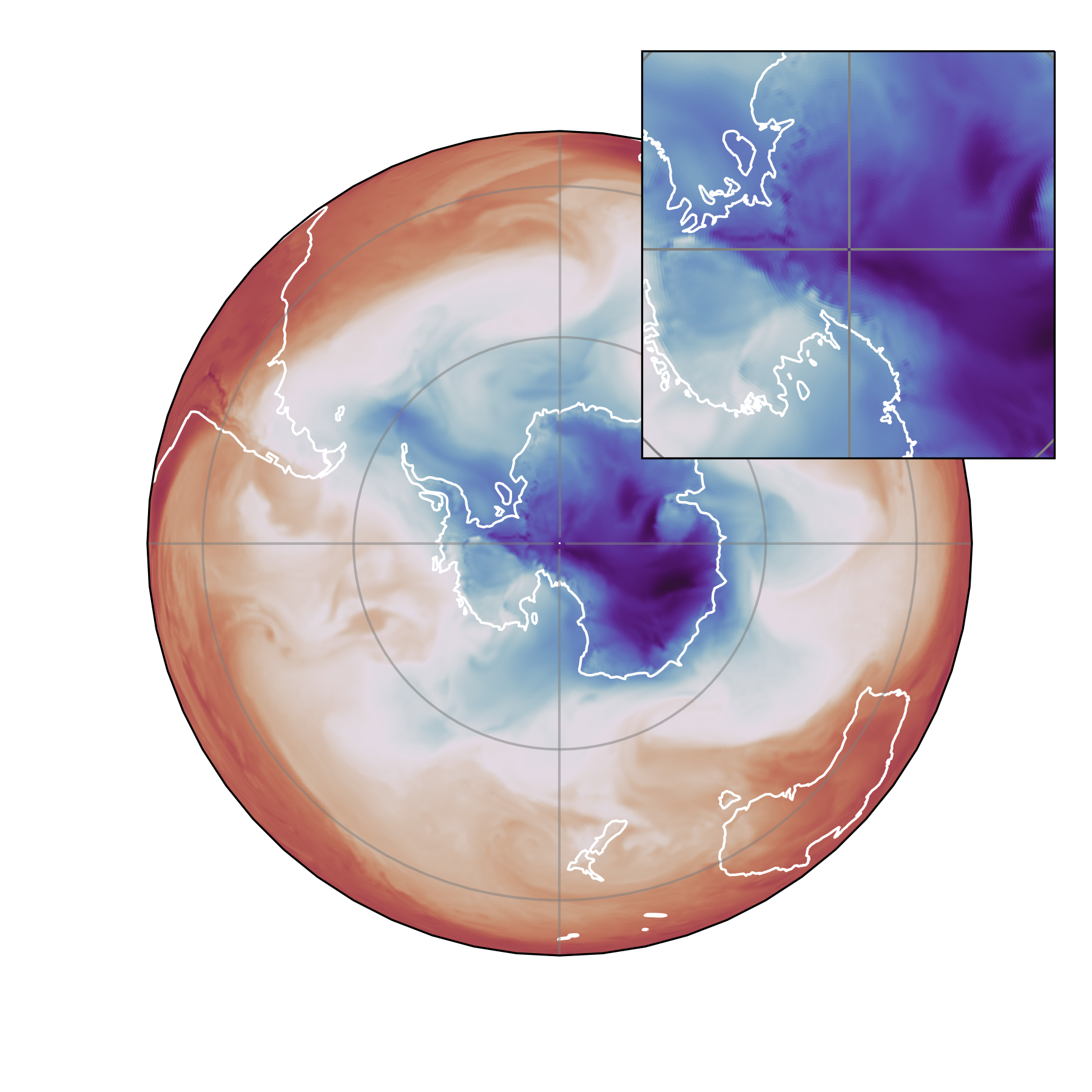}
        \caption{ground truth}
    \end{subfigure}
    \hfill
    \begin{subfigure}{.48\linewidth}
        \includegraphics[width=\linewidth, trim={14px 14px 14px 14px}, clip]{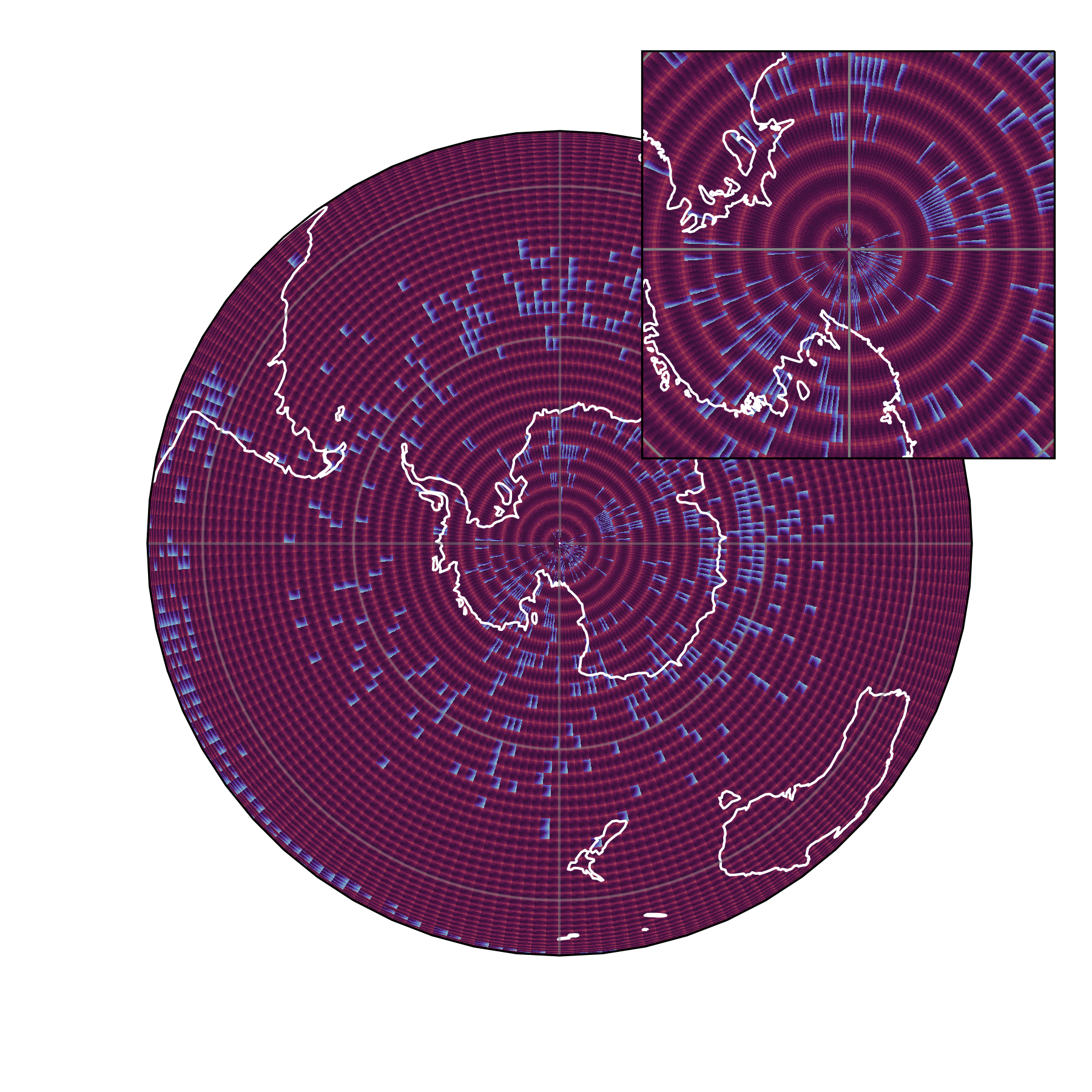}
        \caption{AFNO}
    \end{subfigure}
    \\
    \begin{subfigure}{.48\linewidth}
        \includegraphics[width=\linewidth, trim={14px 14px 14px 14px}, clip]{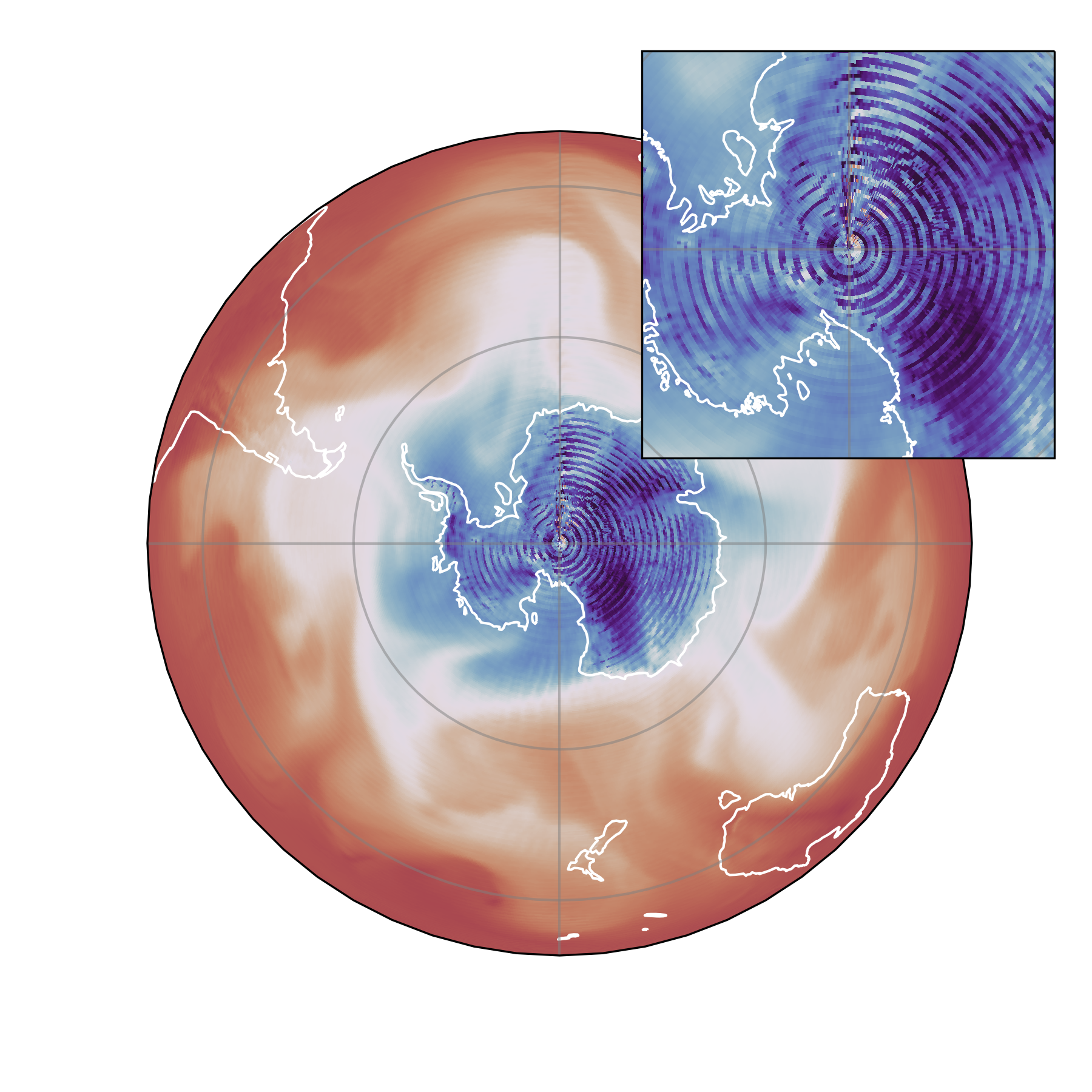}
        \caption{FNO, non-linear}
    \end{subfigure}
    \hfill
    \begin{subfigure}{.48\linewidth}
        \centering
        \includegraphics[width=\linewidth, trim={14px 14px 14px 14px}, clip]{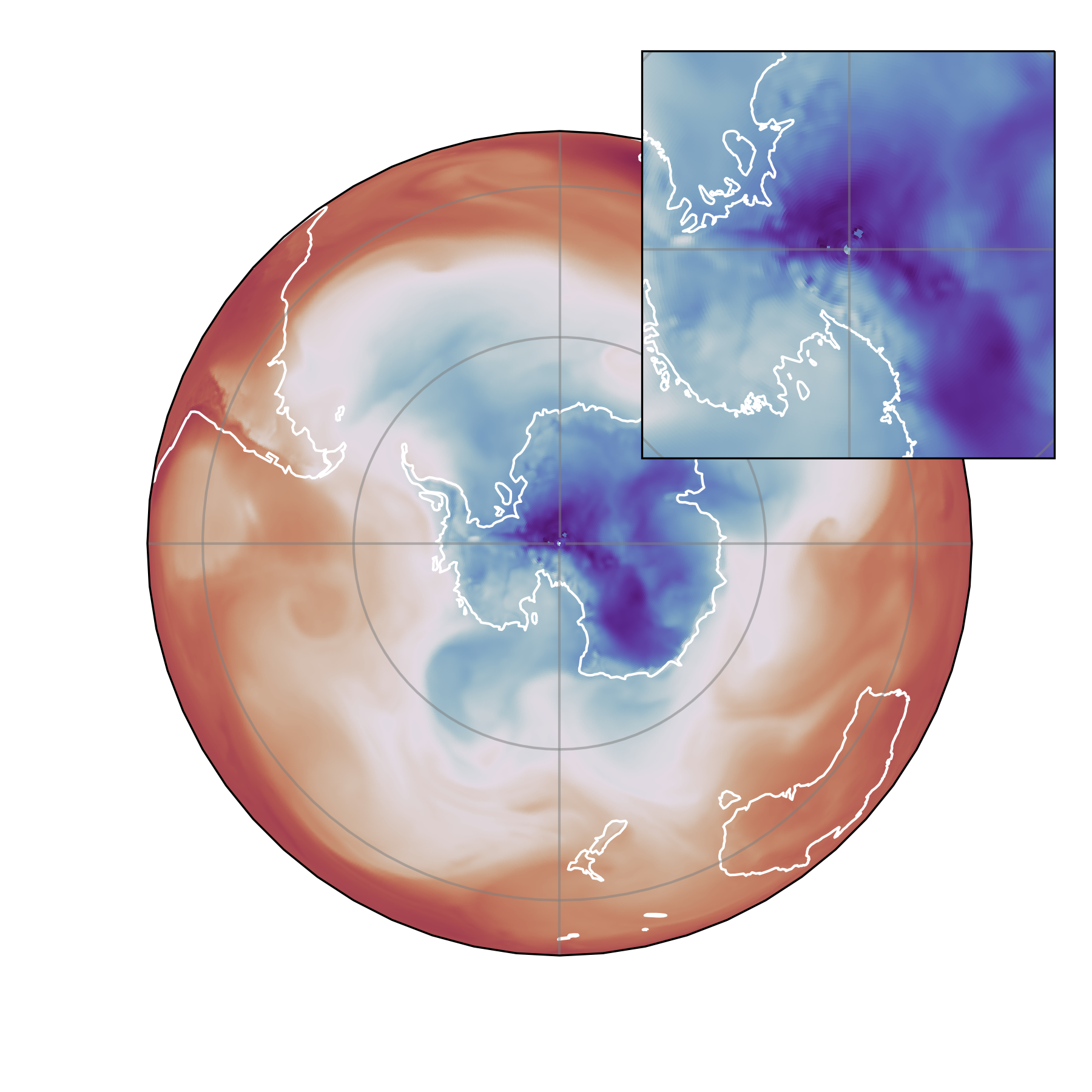}
        \caption{SFNO, linear}
    \end{subfigure}
    \caption{Qualitative comparison of temperature predictions (\texttt{t850}) over Antarctica at 4380h (730 autoregressive steps). The SFNO shows no visible artifacts even after six-month-long rollouts. Models which use FFT and do not incorporate spherical geometry are not stable for long rollouts compared to SFNO.  The AFNO model breaks down early and shows large visible artifacts everywhere. In the non-linear FNO model, artifacts are less pronounced but increase in magnitude with time and towards the poles.}
    \label{fig:era5_t850_southpole}
\end{figure}

Climate change is arguably one of the greatest challenges facing humanity today. Modeling Earth's complex weather and climate accurately, and in a computationally efficient manner, has massive implications for science and society across the enterprise of climate prediction, mitigation, and adaptation. 

Weather and climate modeling has traditionally relied on principled physics- and process-based numerical simulations that solve the partial differential equations (PDEs) governing the fluid dynamics, thermodynamics, and other physics of the Earth system. These equations are discretized and solved on a grid, but the wide range of spatial and temporal scales, as well as complex nonlinear interactions across these scales, necessitate fine grids and high resolution making these computations extremely expensive.

Machine learning (ML) provides alternative approaches to modeling weather and climate, and more generally, spatio-temporal dynamics, by describing the time evolution of the system as a learned transition map between states of the time-discretized physical system exclusively from raw data. While this enables a unified treatment of the full system, the physics is deduced from data alone without imposing the strong inductive bias of the aforementioned physics-based models. Hence, purely data-driven ML-based methods have struggled to faithfully represent the dynamics of spatio-temporal physical systems, especially those with long-range correlations in space and time.

Fourier Neural Operators \cite{FNO} and their variants \cite{AFNO, wen2022u, Rahman2022, kovachki2023neural, kossaifi2023multigrid} possess the advantage of learning mappings between function spaces, which act globally on the entire domain. In contrast, standard neural networks, such as convolutional neural networks and vision transformers, learn on a fixed discretized grid and fail to capture the fine scales of multi-scale systems. Further, the highly optimized Fast Fourier Transform (FFT) in FNO allows for modeling global, long-range interactions in quasi-linear time. Thus, in addition to being discretization invariant, FNO is also computationally more efficient compared to standard vision transformers, which have quadratic time complexity, and more effective in capturing global dependencies compared to convolutional neural networks, graph neural networks, and other local models~\cite{mccormick1987multigrid, falk2019u}.  However, a drawback of FNOs is that FFT is defined on a Euclidean domain leading to incorrect identification of the north and south pole as well as incorrect longitudinal periodicity on the two-sphere $S^2$.

\textbf{Our approach:} We extend the FNO approach to respect the geometry of $S^2$ and its associated symmetries. To do so, we utilize the generalized Fourier transform, which projects functions defined on compact Riemannian manifolds $\mathcal{M}$ onto eigenfunctions of the corresponding Laplace-Beltrami operator, which form an orthogonal basis of the Hilbert space $L^2(\mathcal{M})$.
In the spherical setting $\mathcal M = S^2$, these are the Spherical Harmonics, and the Spherical Harmonic Transform (SHT) generalizes the Fourier transform \cite{Driscoll1994}.
We formulate the SFNO in a manner consistent with the convolution theorem on the sphere. The resulting operator satisfies equivariance properties, such that rotating the input to the operator commutes with the operator itself. Translational or rotational equivariance also motivates the formulation of physical theories, as we do not expect the physical laws to change with a changed frame of reference. This rotational equivariance is, therefore, a strong inductive bias to the learned operator.

In addition, the SFNO retains the favorable properties of FNOs. The learned operators are grid-invariant, since the operations outside of the Fourier/Spherical Harmonic layers act point-wise on the spatial domain. This allows the model to be re-trained at different configurations and resolutions, and even to change the resolution at inference time (i.e., zero-shot super-resolution or interpolation).

The proposed method is applied to the Earth Reanalysis 5 dataset (ERA5) \cite{hersbach2020era5}, one of the best estimates of Earth's historical weather and climate over the period 1950-present, and to the rotating Spherical Shallow Water Equations (SWE), which are commonly used to model geophysical fluid dynamical phenomena \cite{Nair2005, Bonev2018}. In addition to accurate predictions, our method leads to greatly increased long-term stability, with autoregressive inference remaining stable for over one year (1,460 steps) as opposed to 25 days (100 steps) with a comparable, FFT-based method (FNO). As each autoregressive step takes around 500ms on an NVIDIA A6000 GPU, i.e., less than 13 minutes for a year-long simulation, these developments open the door to long-range ensemble inference and uncertainty quantification, well beyond weather timescales to subseasonal-to-seasonal (S2S) prediction and potentially, towards climate prediction.

\textbf{Key contributions}:
\begin{itemize}
    \item \textbf{A novel SFNO equivariant architecture} for modeling nonlinear chaotic dynamical systems on the sphere.
    \item Theoretical extension of FNOs to spherical geometry, with the desirable properties of \textbf{equivariance} while retaining \textbf{grid-invariance}.
    \item  A demonstration of exceptional 
 long-term \textbf{stability} of the auto-regressive map $F$, on year-long rollouts, while observing plausible dynamics of the predicted physical quantities.
    \item  \texttt{torch-harmonics}, an efficient and differentiable implementation of the spherical harmonics transform in 
\texttt{PyTorch}\footnote{\texttt{torch-harmonics} and our implementation of the SFNO are available to the public at \url{https://github.com/NVIDIA/torch-harmonics}}, enabling \textbf{scalable model parallelism}s.  
\end{itemize}

The rest of the paper is organized as follows.  Related work is discussed in Section \ref{sec:context_and_related_work}. Section \ref{sec:SFNO} presents the theoretical foundations used during SFNO design. The network topology design is then laid out in Section \ref{sec:equivariant_architecture}. Section \ref{sec:numerical_experiments} documents the experimental setup and training infrastructure. Section \ref{sec:conclusion} concludes the paper.

\section{Related Work and Broader Context}
\label{sec:context_and_related_work}

ML-based weather and climate modeling is less than five years old, yet has seen a massive surge in capability thanks to advancements in deep learning \cite{Yuval2020, Pathak2022, Bi2022, Lam2022}. Relatively simple architectures like CNN-based encoder-decoder architectures \cite{scher2019weather} have been applied to two-dimensional latitude-longitude grids. 
\citet{weyn2020improving, weyn2021sub} noticed the limitations of this representation when treating the poles and proposed an extension based on the equiangular gnomonic cubed sphere. 
\citet{Pathak2022} combined a Fourier-transform-based token-mixing scheme \cite{AFNO} with a vision transformer (ViT)
backbone \cite{dosovitskiy2020image} and developed FourCastNet (FCN) to model long-range dependencies.
More recently, \citet{Bi2022} developed another ViT-based approach they call Pangu-Weather, which relies on an encoder-decoder architecture and demonstrated that ML-based methods are able to compete with and even outperform classical physics-based numerical methods for numerical weather prediction (NWP). Most recently, GraphCast, another approach based on graph
neural networks \cite{Lam2022} developed the first approach to work on a spherical geometry via a multi-scale mesh representation and demonstrated superior predictive performance on weather forecasting. Except \citet{weyn2021sub}, none of the above-mentioned models were stable beyond a few weeks of the autoregressive rollout. \citet{weyn2021sub}, however, were limited in their predictive performance and operated on a resolution that was too low to provide useful information on the regional and local scales at which extreme weather impacts society. Therefore, there remains a strong need for ML-based high-resolution stable long-term forecasts of atmospheric phenomena. 

Concepts such as equivariance, symmetry, and learned representations in spectral domains exist in many different contexts. An exhaustive review of these is beyond the scope of this paper and we refer the reader to the excellent overview on geometric deep learning by \citet{Bronstein2021}. 

\textbf{Introducing inductive biases} to regularize machine learning problems have a long history in computer vision and machine learning.
The inductive bias of convolutional layers to achieve translation invariance resulted in a decade of the predominance of convolutional neural networks (CNNs) \cite{lecun95convolutional}.
Nowadays, ViTs employ self-attention as the mechanism to exploit the Symmetric group $G=S_n$ of permutations acting on $n$ input tokens as a less restrictive, yet, effective symmetry to relax the inductive bias of translational equivariance in CNNs \cite{pmlr-v139-bertasius21a}. \textbf{Equivariance} has gained significant attention in the past few years, including group equivariant CNNs \cite{Cohen2016} and their spherical counterparts \cite{Cohen2018, Ocampo2022}. 

We provide an equivariant extension of Fourier Neural Operators to the sphere, based on a convolution theorem on the sphere. 

\section{Background}
\label{sec:SFNO}

\subsection{Problem Setting}

We model the Earth's atmosphere as a dynamical system, where its state at time $t$ and position $x$ is represented by a $N$-dimensional vector $\vb{u}(x, t) : S^2 \times [0, T) \rightarrow \mathbb{R}^{N}$.
Each of these variables represents a physical quantity of interest, such as pressure, wind velocity, humidity, etc. The system is observed only partially, meaning that there might be hidden variables that influence the dynamics, which are not contained in this description. Moreover, the system is observed only at discrete times $t_n \in \{t_0, t_1, \dots\}$, and at discrete positions $\vb{x}_i \in \{\vb{x}_1, \vb{x}_2, \dots, \vb{x}_N\}$. For simplicity, we assume that there exists a deterministic mapping 
\begin{equation}
  \vb{u}_{n+1} = F[\vb{u}_{n}, t_n],
\end{equation}
which maps the discrete state vector $\vb{u}_{n}$ at time $t_n$ to the state $\vb{u}_{n+1}$. Our aim is to learn this map from data. This is in contrast to the classical approach using PDEs, where this map is typically inferred from first principles by discretization and numerical integration. Carefully crafted numerical schemes often conserve certain symmetry properties of the underlying PDE, which motivate the goal of formulating an equivariant ML method.

\subsection{Fourier Neural Operators}

Our construction of equivariant mappings between function spaces on the sphere is an extension of the FNO framework. FNO learns a resolution-independent representation using a global convolution kernel $\kappa$
\begin{align}
    \mathcal K\bigl[u\bigr](x) &= \int_\mathcal{M} \kappa (x-y) \cdot u(y) \,\mathrm d y &\  \forall x \in \mathcal{M} \ , \label{eq:convolution}
\intertext{which can be rewritten in terms of the convolution theorem}
\mathcal K\bigl[u\bigr](x) &= \mathcal F^{-1}\bigl[ \mathcal F[\kappa] \cdot \mathcal F[u] \bigr](x) &\  \forall x \in \mathcal{M} \ .
\label{eq:convolution_theorem}
\end{align}
The continuous Fourier transform $\mathcal{F}$ can be expressed as the DFT when sampling the finite domain $\mathcal M$ using a uniform grid. Hence, FNOs allow for expressing long-range dependencies using global convolutions while employing an efficient implementation of the FFT, guaranteeing log-linear asymptotic time complexity. 

The Spherical FNO (SFNO) layer extends this approach to adapt the Fourier transform while respecting the symmetry of the underlying manifold.

\subsection{Incorporating Symmetries on General Manifolds}

Symmetries are of crucial importance for the analysis of physical systems and historically played a seminal role in the theoretical formulation of natural phenomena long before the first experimental evidence was gathered for their existence. Group theory-driven breakthroughs include the systematic classification of lattices in crystallography \cite{Sloan}, the postulation of novel particles such as quarks \cite{Gell-Mann:1961omu}, and the standard model of particle physics \cite{Workman:2022ynf}. Respecting the intrinsic geometry by means of Lie group actions $\Phi: G \times \mathcal M \rightarrow \mathcal M\,,\, (g, x) \mapsto \Phi(g, x)$ acting on the configuration space manifold $\mathcal M$ of spatial coordinates allows for: (i) the identification of conserved quantities by Noether's theorem; and (ii) the elimination of redundant variables, which usually leads to a decoupling or simplification of the associated equations of motion.

In ML, symmetries act as hard constraints limiting the search space to physically meaningful sub-manifolds/orbifolds, which may speed up optimization as well as improve the robustness of the solution through explicit regularization. A prominent example is convolutional layers \cite{fukushima:neocognitronbc}, which constrain the space of trainable parameters to the inner product of a filter $k$ being translated along the orbit parametrized by the group action $\Phi(x, y) := y - x$ of the translation group $G=(\mathbb R, +)$. This leads to the well-known expression for convolution.\footnote{While convolutional layers typically implement cross-correlation, we choose to discuss convolutions for the sake of simplicity.}
\begin{alignat}{100}
    \bigl(\kappa \star u\bigr) (x)  &= \int  \kappa\bigl(\Phi(x, y)\bigr) \cdot u(y)\ \mathrm d y \nonumber \\ &= \int  \kappa(y-x) \cdot u(y) \ \mathrm d y  \quad  \forall x \in \mathbb{R}.\ \label{eq:crosscorrelation}
\end{alignat}
 While the Abelian translation group $G=(\mathbb R^n, +)$ naturally acts on flat $\mathbb R^n$ via plain vector addition, we cannot globally define a convolution operation on curved manifolds by means of axis-oriented translations.

Hence, for physical systems on general manifolds $\mathcal M$, we extend the Fourier transform as well as the operations acting on the feature representations to respect their geometric constraints. On compact Riemannian manifolds, the generalized Fourier transform can be understood as the map that decomposes functions in terms of the eigenfunctions of the Laplace-Beltrami associated with the underlying manifold $\mathcal M$. In general, this view neglects the group structure if $\mathcal M$ is a Lie group.

For all practical purposes, we are interested in dynamical systems formulated on the sphere $\mathcal M = S^2 \subset \mathbb R^3$. The canonical choice on the sphere is rotational group actions stemming from the set of orientation-preserving isometries of the scalar product in $\mathbb R^3$, namely the special orthogonal group $SO(3)$. We consider the Hilbert space of square-integrable functions with respect to the Lebesgue measure defined on the sphere $L^2(S^2)$. We define the (partial map of a) group action for an arbitrary but fixed rotation $\Phi_R: L^2(S^2) \mapsto L^2(S^2)$ onto a square-integrable function $u \in L^2(S^2)$ as the result of transforming the spatial coordinates with the corresponding passive rotation $R^{-1} \in SO(3)$.
\begin{align}
    \Phi_R[u](x) := u(R^{-1} x)\quad \forall R \in SO(3) 
\end{align}
In other words, in order to rotate an image on the sphere, we rotate its pre-image in the opposite direction. This is analogous to the planar case where a function is shifted along the coordinate axes using the translation group $G=(\mathbb R^n, +)$. In the following, we are interested in mappings $F: L^2(S^2) \rightarrow L^2(S^2)$, that are compatible with the aforementioned group actions on the Hilbert space.
We call a transformation $F$ equivariant with respect to two $SO(3)$ group actions $\Phi_R, \Psi_R$ both acting onto $L^2(S^2)$, iff 
\begin{align}
\Psi_R\bigl[F[u]\bigr]&= F\bigl[\Phi_R[u]\bigr] \quad \forall R \in SO(3)    
\end{align}
Simply put, it does not matter whether one initially rotates the coordinate system followed by a subsequent transformation on the function space or vice versa. For simplicity, we stretch the notation and drop the explicit group actions $\Phi, \Psi$ and identify them with the multiplication of group elements. Equi\-variance then reads: $F \circ R = R \circ F$ for all $ R \in SO(3)$.

\subsection{Fourier Transform on the Sphere}

In flat geometry, the Fourier transformation is a change of basis by projecting a function $u \in L^2(\mathbb R^n)$ onto planar waves $b_k(x)=\exp\bigl(i\langle k , x \rangle\bigr)$ effectively encoding translation equivariance. In a general setting, the set of basis functions can be obtained from the eigenfunctions of the Laplace-Beltrami operator. In the following, we will provide an explicit construction for spherical geometry.

On $S^2$, these are the spherical harmonics  \cite{abramowitz1964handbook} defined as
\begin{equation}
    \label{eq:spherical_harmonics}
    Y_l^m(\theta, \varphi):=(-1)^m c_l^m P_l^m(\cos \theta) e^{\mathrm{i} m \varphi},
\end{equation}
which form an orthogonal basis of $L^2(S^2)$. $P^m_l$ and $c^m_l$ denote the associated Legendre polynomials and normalization factors\footnote{For a detailed introduction of the spherical harmonics see Appendix \ref{sec:spherical_harmonics}.}. More importantly, among all the possible bases of $L^2(S^2)$, the spherical harmonics uniquely exploit the symmetries of the sphere. Applying rotations $\Psi_R$ to $Y^m_l$ of degree $l$ will result in a linear combination of $Y^m_l$ with the same degree l and $-l \leq m \leq l$.

On the sphere, the corresponding Fourier transformation is the decomposition of the function space $L^2(S^2)$ into minimal subspaces invariant under all the possible rotations in $SO(3)$. The decomposition onto the basis then reads
\begin{alignat}{100}\label{eq:sh-transform}
& u(\theta, \phi) &&= \sum_{l \in \mathbb{N}} \sum_{|m| \leq l} \hat{u}(l, m) Y_l^m(\theta, \phi) \nonumber \\
& \hat{u}(l, m)&&=\int_{S^2} \overline{Y_l^m} \cdot u \; \dd \Omega \ ,
\end{alignat}
where $\mathrm d \Omega = \sin \theta\, \mathrm d \theta \,\mathrm d \phi$ is the volume form of the sphere. $\mathcal{F}: u \rightarrow \hat{u}$, maps functions in $u \in L^2(S^2)$ to the harmonic coefficients $\hat{u}(l, m)$. We call this map Spherical Harmonic Transform (SHT), or alternatively, generalized Fourier transform \cite{Driscoll1994}.

The Fourier transform decomposes a Hilbert space into basis functions associated to 
symmetry operations on the underlying manifold. For the translation group $G=(\mathbb R^2, +)$, this basis is given by tensor products of plane waves $b_{kl}(x,y)=\exp(ikx) \otimes \exp(ily) $ since translations
along the axes commute with each other. For the sphere, the situation is more complicated as the generators of rotations do not commute in general. This is reflected by the fact that
the associated spherical basis functions do not factor into a tensor product structure.

\section{Spherical Fourier Neural Operator}
\label{sec:sfno}
\subsection{Convolutions on the Sphere}
We can now generalize the FNO naively by replacing the Fourier transforms in \eqref{eq:convolution_theorem} with the SHT. While the SHT generalizes the Fourier transform on $S^2$, this ignores that \eqref{eq:convolution_theorem} is derived from a convolution theorem, however, and ignores the symmetries inherent to the sphere. Instead, we seek a formulation motivated by a convolution operator to obtain an equivariant formulation. We introduce the spherical convolution 
\begin{equation}\label{eq:convolution_s2}
    (\kappa \star u)(x) = \int_{R \in SO(3)}\kappa(R n) \cdot u(R^{-1} x) \mathrm{d}R, \; \forall x \in S^2.
\end{equation}
This definition applies the rotation $R$ to the northpole $n = (0,0,1)^T$, which makes the function $\kappa \in L^2(S^2)$ compatible with rotations $R \in SO(3)$. This formulation generalizes the usual definition of convolutions, which can be recovered by replacing the north pole $n$ with the origin $0$ and rotations $R$ with translations $T \in (\mathbb{R}^n, +)$.

The spherical convolution \eqref{eq:convolution_s2} admits a convolution theorem of the form
\begin{equation}\label{eq:convolution_theorem_s2}
    \mathcal{F}[\kappa \star u](l, m) = 2 \pi \sqrt{\frac{4 \pi}{2l + 1}}\; \mathcal{F}[u](l, m) \cdot \mathcal{F}[\kappa](l, 0),
\end{equation}
where $\mathcal{F}$ refers to the SHT\footnote{For a detailed discussion of Fourier transforms and convolutions on the sphere, we refer the reader to \cite{Driscoll1994}.}. By replacing the filter weights $\mathcal{F}[\kappa](l, 0)$ with the learned weights $\tilde{\kappa}_\vartheta(l)$, we obtain the Spherical Fourier Neural Layer
\begin{equation}\label{eq:sfno}
\mathcal{K}_\vartheta[u] = \mathcal{F}^{-1}[\tilde{\kappa}_\vartheta \cdot \mathcal{F}[u]],
\end{equation}
which forms the core of the SFNO. More precisely, we have
\begin{equation*}
\mathcal{F}[\mathcal{K}_\vartheta[u]](l,m) = \tilde{\kappa}_\vartheta(l) \cdot \mathcal{F}[u](l,m),
\end{equation*}
as this approach requires only one filter weight per $l$ to be learned. If $L$ is the maximum zonal mode ($0 \leq l \leq L$), this approach requires $L$ learned filters, as opposed to $L(L+1)/2$ for the naive extension of the FNO.



\subsection{SFNO Network Topology Design}
\label{sec:equivariant_architecture}

\begin{figure}[h]
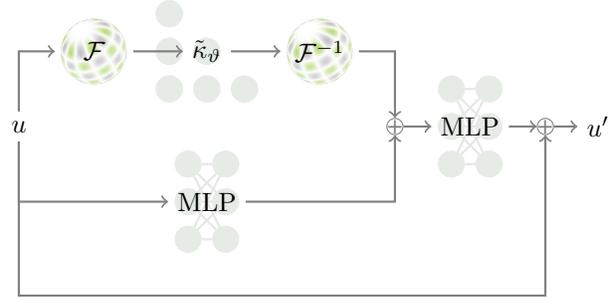

    \centering
    \include{figures/sfno/spherical_fourier_neural_operator.tex}
    \vspace{-10pt}
    \caption{The structure of a single SFNO block. Multi-layer perceptrons (MLPs) act point-wise in the spatial domain and allow for channel mixing. The generalized Fourier transform $\mathcal{F}$ and its inverse $\mathcal{F}^{-1}$ allow for the learning of long-range spatial dependencies. $\kappa$ is a learned filter, which is applied linearly to the frequency components.}
    \label{fig:diagram_sfno}
\end{figure}
\begin{figure}[h]
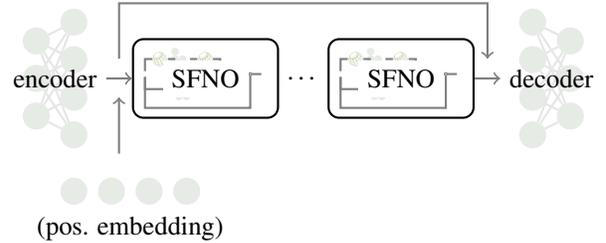

    \centering
    \include{figures/sfno/architecture.tex}
    \vspace{-20pt}
    \caption{Diagram of the overall SFNO architecure. Encoder and decoder MLPs inflate the channel dimension. A learned position embedding is added in cases where position-dependent information should be learned by the network. At the core lie $N$ SFNO blocks, where the first and last blocks perform up- or down-scaling. A skip connection is added for autoregressive maps close to the identity.}
    \label{fig:diagram_architecture}
\end{figure}

\textbf{SFNO block:}
Figure \ref{fig:diagram_sfno} depicts the layout of a single SFNO block. At the core lies the Fourier layer \eqref{eq:sfno}, which allows the network to efficiently learn global correlations. The formulation in terms of a spherical convolution makes the Fourier layer linear and equivariant w.r.t. $SO(3)$. In case of vector-valued inputs $u$, $\tilde{\kappa}_\vartheta(l)$ is replaced by a complex-valued, learned matrix for each $l$ and the multiplication in \eqref{eq:sfno} is replaced by a matrix-vector product.

Alternatively, we propose the use of a complex-valued neural network, which acts "frequency-wise" in the Fourier domain. This approach is an extension of the AFNO architecture \cite{AFNO}. The non-linear approach is not equivariant w.r.t. $SO(3)$, however. The study of complex-valued neural networks is still in its infancy, and they are not as well understood as dense, ReLU networks \cite{Voigtlaender2020}. For this reason, we chose one possible extension of the ReLU activation, which applies it only to its real component: $x + iy \rightarrow \mathrm{ReLU}(x) + iy$.

\textbf{Point-wise nonlinearities:}
The remaining components of our networks are equally chosen with equivariance in mind. Any operation acting point-wise on $u$ is trivially equivariant in the continuous setting. This is not the case in the discrete setting, however, as non-linear functions may introduce arbitrary frequency components, which makes them only approximately equivariant \cite{Karras2021}. In principle, this can be remedied by applying non-linearities in the frequency domain \cite{Poulenard2021}. We refrain from doing so as this leads to large performance overheads and limit ourselves to equivariance in the continuous limit.

\textbf{Equivariant rescaling:}
To reduce the memory footprint of our models, we employ down- and up-scaling and keep a lower-resolution hidden state in our models. Adaptive FNOs, as used in FCN, realize this via patching in the encoding layer and pixel-shuffling in the decoding network \cite{AFNO, Pathak2022}, which is inherently not equivariant on the sphere. We propose using the SFNO block directly to perform up- and down-scaling. This is achieved by truncating the frequencies in the forward transform $\mathcal{F}$ and evaluating the inverse $\mathcal{F}^{-1}$ at a higher resolution when up-scaling. This is also applied to the skip connections to obtain the residual in the desired up- or down-scaled resolution.

Figure \ref{fig:diagram_architecture} depicts the structure of the overall network $F_\vartheta: u_n \rightarrow u_{n+1}$. It consists of three main parts: An encoder network, multiple spherical FNO blocks, and a decoder network. To maintain equivariance properties, the encoder and decoder networks are also point-wise MLPs with a single hidden layer and GELU activations \cite{GELU}. These layers inflate and deflate the channel dimension to the embedding dimension, which remains constant throughout the network. As the autoregressive map $F$ is close to identity, we add a large skip connection that feeds the output of the encoder to the decoder, skipping the SFNO blocks. To model spatial dependencies, a learned position embedding is added after the decoder layer. This can be regarded as an additional, learned input, and is, therefore, equivariant as well. As previously noted, the first and last SFNO blocks are used to perform up- and down-sampling. Except for the last layer, instance norm \cite{ulyanov2016instance} is applied after each MLP to normalize the inputs. We choose instance norm over other normalization methods, as it applies the same operation globally, making it equally equivariant.

We remark that the obtained model is grid-invariant, as all learned operations save for the positional embedding do not depend on the grid of the input data. This allows the model to be applied on arbitrary grids and resolutions as long as the SHT $\mathcal{F}$ and its inverse $\mathcal{F}^{-1}$ can be computed on it. Moreover, the position embedding can be parametrized in terms of spherical harmonics to facilitate the grid-invariance of the architecture.

\textbf{Differentiable Spherical Harmonic Transform:} To enable our method, we implement \texttt{torch-harmonics},
a library for differentiable Spherical Harmonics written in \texttt{PyTorch} \cite{Paszke2019}. We choose the ``direct'' algorithm \cite{Schaeffer2013}, as it has advantages over algorithms for the implementation on GPUs. Details on the implementation are provided in Section \ref{sec:differentiable_sht}.

\begin{figure*}[tb]
    \centering
    \begin{subfigure}{.24\linewidth}
        \centering
        \raisebox{18mm}{
        \begin{subfigure}{\linewidth}
            \includegraphics[width=\linewidth, trim={20px 20px 20px 20px}, clip]{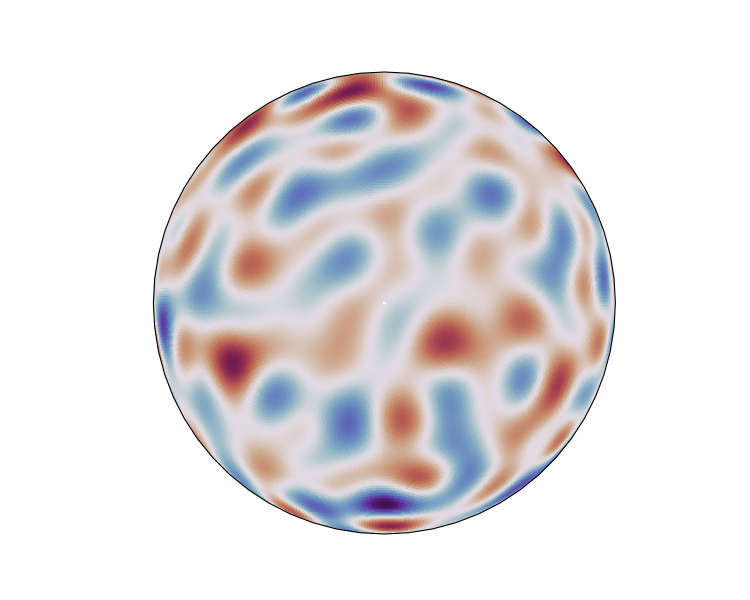}
            \caption{initial condition, $t=0h$}
        \end{subfigure}}
    \end{subfigure}
    \hfill
    \begin{subfigure}{.24\linewidth}
        \centering
        \begin{subfigure}{\linewidth}
            \includegraphics[width=\linewidth, trim={20px 20px 20px 20px}, clip]{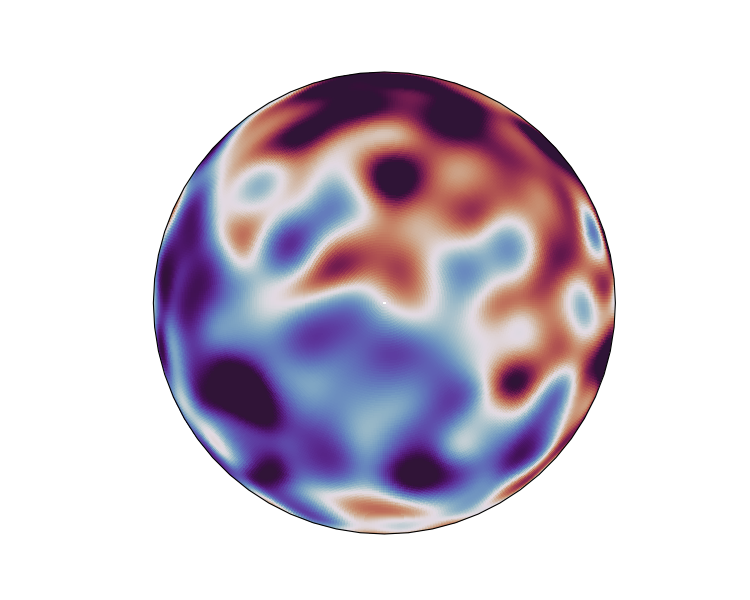}
            \caption{ground truth, $t=5h$}
        \end{subfigure}
        \\
        \begin{subfigure}{\linewidth}
            \includegraphics[width=\linewidth, trim={20px 20px 20px 20px}, clip]{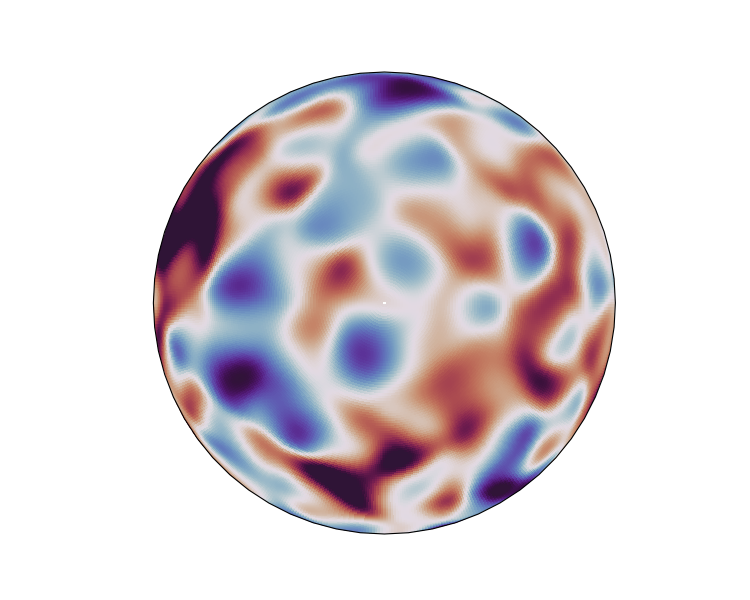}
            \caption{ground truth, $t=10h$}
        \end{subfigure}
    \end{subfigure}
    \hfill
    \begin{subfigure}{.24\linewidth}
        \centering
        \begin{subfigure}{\linewidth}
            \includegraphics[width=\linewidth, trim={20px 20px 20px 20px}, clip]{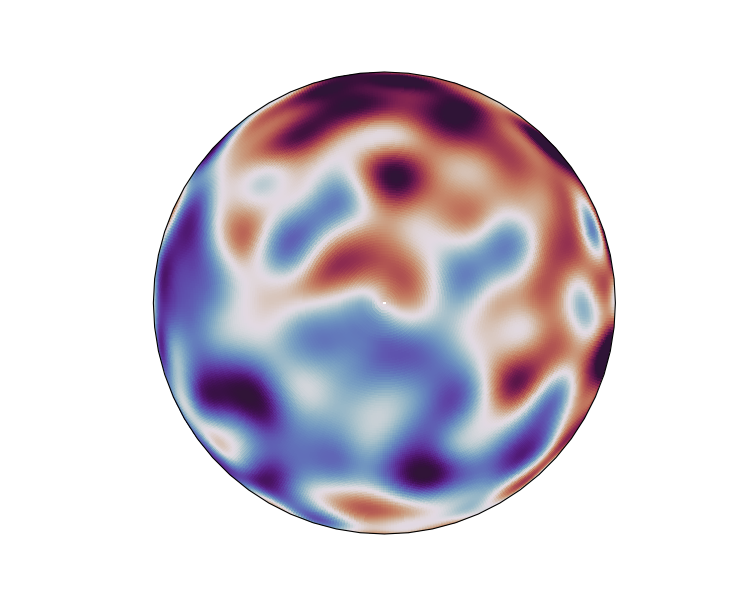}
            \caption{SFNO, $t=5h$}
        \end{subfigure}
        \\
        \begin{subfigure}{\linewidth}
            \includegraphics[width=\linewidth, trim={20px 20px 20px 20px}, clip]{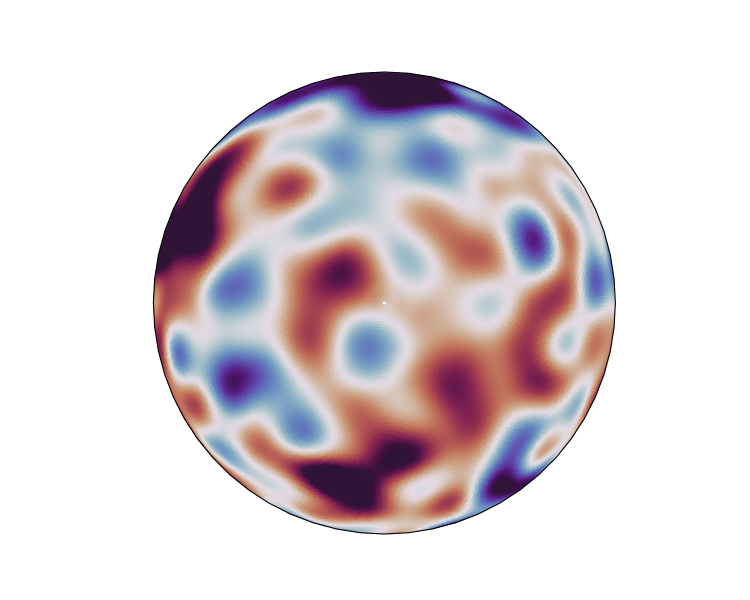}
            \caption{SFNO, $t=10h$}
        \end{subfigure}
    \end{subfigure}
    \hfill
    \begin{subfigure}{.24\linewidth}
        \centering
        \begin{subfigure}{\linewidth}
            \includegraphics[width=\linewidth, trim={20px 20px 20px 20px}, clip]{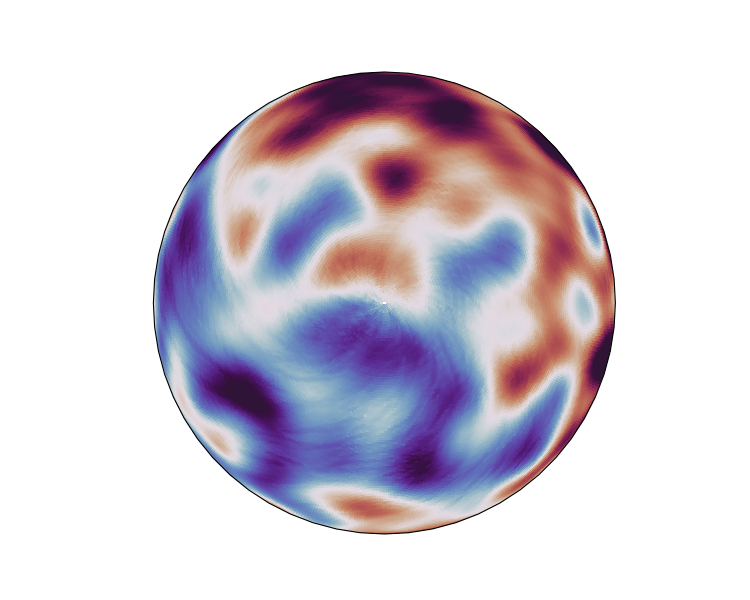}
            \caption{FNO, $t=5h$}
        \end{subfigure}
        \\
        \begin{subfigure}{\linewidth}
            \includegraphics[width=\linewidth, trim={20px 20px 20px 20px}, clip]{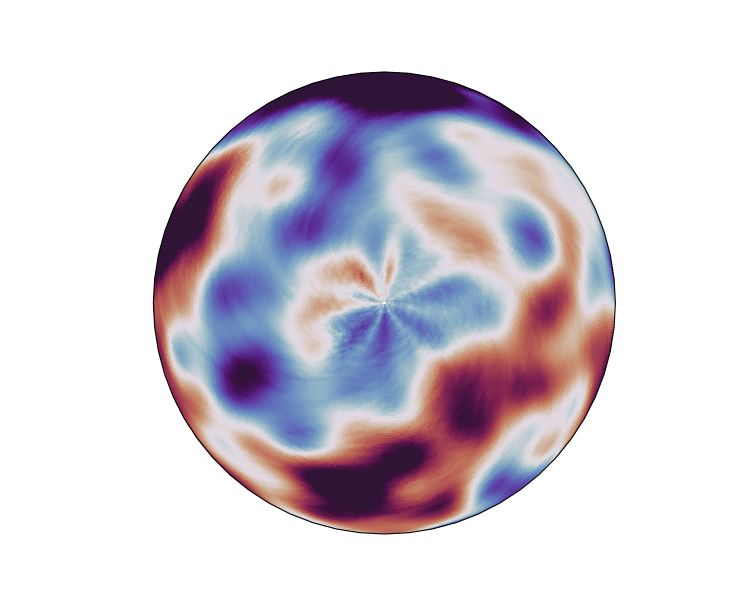}
            \caption{FNO, $t=10h$}
        \end{subfigure}
    \end{subfigure}
    \caption{Solutions to the Shallow Water Equations on the rotating Sphere predicted by SFNO and FNO architectures in comparison to the ground truth solution computed using a classical spectral solver. Plots depict the geopotential height at 5 and 10 hours, corresponding to 5 and 10 autoregressive steps respectively. The view is centered on the south pole to highlight artifacts arising due to the non-geometrical treatment of the sphere.}
    \label{fig:swe_examples}
\end{figure*}

\begin{table*}[tb]
    \caption{Auto-regressive results applying the Shallow Water Equations on the rotating sphere at a spatial resolution of 256 $\times$ 512 and a temporal resolution of 1 hour. Losses are relative and reported at 1h and 10h respectively, which correspond to 1 and 10 auto-regressive steps. Losses for the ML models are computed w.r.t. the validation data generated by the classical solver. To put them into perspective, the error of the classical method, and therefore the training data is listed in the bottom row. These errors are estimated using a high-fidelity solution computed at four times the resolution of the training data.}
    \label{tab:swe_results}
    \begin{center}
    \begin{small}
    \begin{sc}
    \addtolength{\tabcolsep}{2pt}
    \begin{tabular}{lcccccc}
    \toprule
    \multirow{2}{*}[-1pt]{Model} & \multicolumn{3}{c}{Parameters} & \multicolumn{2}{c}{$L^2$ Loss} & \multirow{2}{*}[-1pt]{eval time}\\
    \cmidrule(lr){2-4} \cmidrule(lr){5-6}
     & \scriptsize{Layers} & \scriptsize{Embed. Dimension} & \scriptsize{Parameter Count}  & \scriptsize{at 1h (1 step)} & \scriptsize{at 10h (10 steps)} &  \\
    \midrule
    U-Net            & 20 & - & $3.104 \cdot 10^7$ & $2.961 \cdot 10^{-3}$ & $1.462 \cdot 10^{-1}$ & 0.011s \\
    \midrule
    FNO, linear     & 4 & 256 & $4.998 \cdot 10^7$ & $8.280 \cdot 10^{-4}$ & $9.958 \cdot 10^{-3}$ & 0.156s \\
    FNO, non-linear & 4 & 256 & $3.920 \cdot 10^7$ & $8.298 \cdot 10^{-4}$ & $9.784 \cdot 10^{-3}$ & 0.212s \\
    \midrule
    SFNO, linear    & 4 & 256 & $3.518 \cdot 10^7$ & $7.741 \cdot 10^{-4}$ & $7.239 \cdot 10^{-3}$ & 0.218s \\
    SFNO, non-linear & 4 & 256 & $3.920 \cdot 10^7$ & $7.673 \cdot 10^{-4}$ & $1.558 \cdot 10^{-2}$ & 0.321s \\
    \midrule
    Classical Solver & - & -   & -                  & $1.891 \cdot 10^{-2}$ & $3.570 \cdot 10^{-2}$ & 1.299s \\
    \bottomrule
    \end{tabular}
    \addtolength{\tabcolsep}{-2pt}
    \end{sc}
    \end{small}
    \end{center}
\end{table*}

\section{Numerical Experiments}

\label{sec:numerical_experiments}

\subsection{Spherical Shallow Water Equations}
\label{sec:shallow_water}

The SWE on the rotating sphere \eqref{eq:swe} are a system of non-linear hyperbolic PDEs modeling the dynamics of a thin fluid layer in the limit where the fluid depth becomes negligible w.r.t. the characteristic wavelength. Thus, they are well-suited to model planetary fluid phenomena such as atmospheric dynamics, tsunami propagation, and tidal flows \cite{Nair2005, Bonev2018}.

We train our models on the SWE by generating random data on the fly using a classical, spectral solver \cite{Giraldo2001}.  We use the data to train four models: two SHT-based and two FFT-based FNOs with either non-linear or linear maps in the frequency domain. Models are trained with the following hyperparameters: 4 (S)FNO blocks, a down-scaling factor of 3, and embedding dimensions of 256. The latter was chosen to make model sizes roughly equal in terms of the number of trainable parameters. The models are trained for 20 epochs using a single autoregressive step. Each time step corresponds to a single hour in the system, which requires the classical solver 150 time steps to compute. In the second stage, two autoregressive steps are used during training to fine-tune the model and improve stability. Details regarding the data generation and training of these models are outlined in Section \ref{app:swe_experiments} in the Appendix.

Figure \ref{fig:swe_examples} shows a comparison of the (linear) SFNO model to ground truth data computed by the classical solver. Detailed numerical results and an overview of the trained models are reported in Table \ref{tab:swe_results}. 

\textbf{Discussion} While both FFT and SHT-based models achieve roughly similar accuracies, it is worth noting that the SFNO models have fewer parameters as the SHT has half as many frequencies as the FFT at the same resolution. Moreover, Figure \ref{fig:swe_examples} depicts autoregressive rollouts of 5 and 10 hours for the linear FFT and SHT-based models. It is evident that the FFT-based approach leads to artifacts visible at the poles, which increase in severity for longer rollouts.

Training the individual models takes 240s on a single NVIDIA A6000. This is mainly bottlenecked by the numerical solver, which requires 150 time steps in order to generate a single sample 1h in advance.

\subsection{Weather Prediction / Atmospheric Dynamics}

\begin{figure}[ht!]
    \centering
    \begin{subfigure}{\linewidth}
        \centering
        \includegraphics[width=0.96\linewidth, trim={100px 40px 100px 40px}, clip]{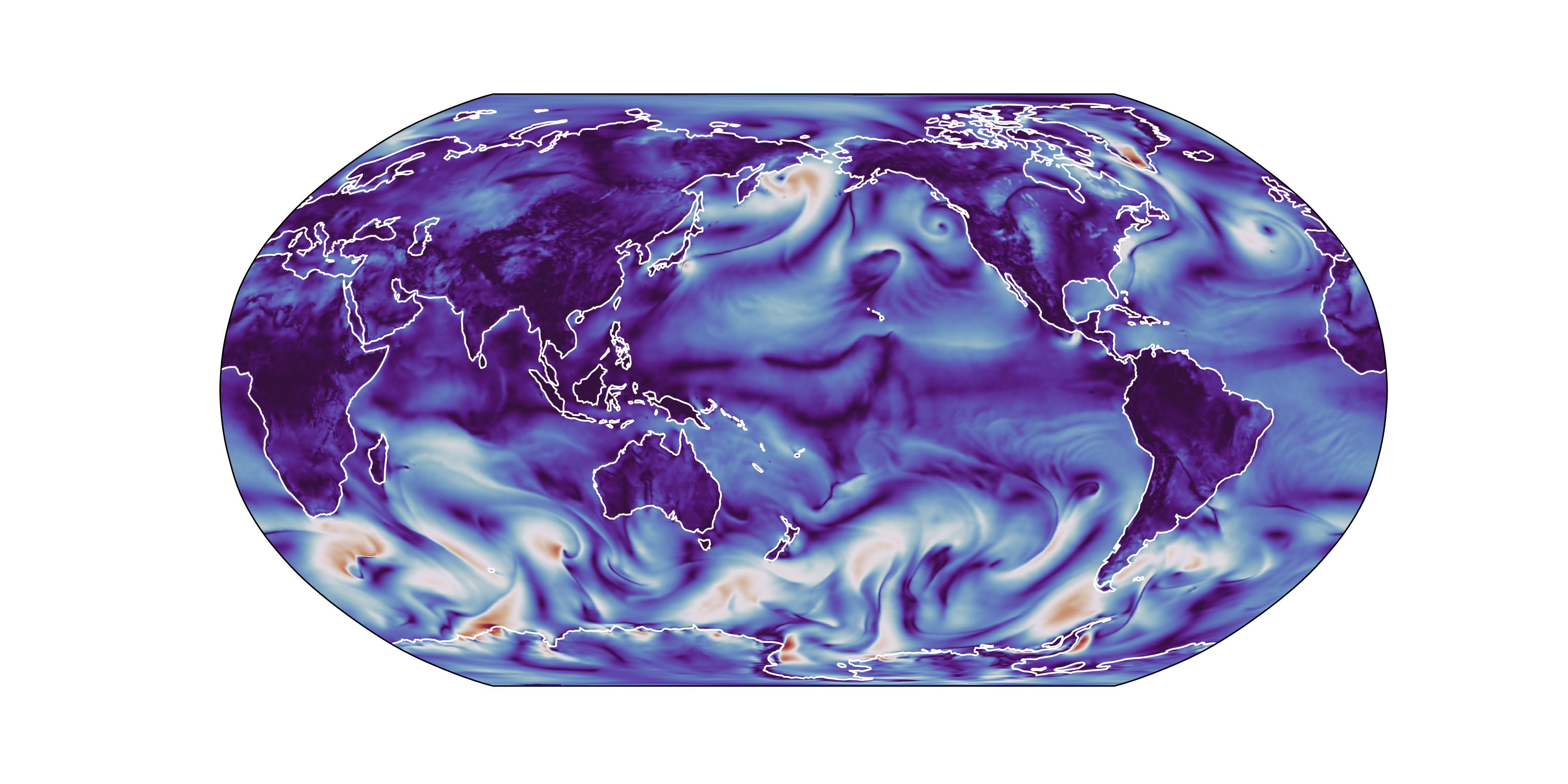}
        \caption{Spherical FNO using the SHT}
        \label{fig:comparison_600h_rollout_sht}
    \end{subfigure}
    \\
    \begin{subfigure}{\linewidth}
        \centering
        \includegraphics[width=0.96\linewidth, trim={100px 40px 100px 40px}, clip]{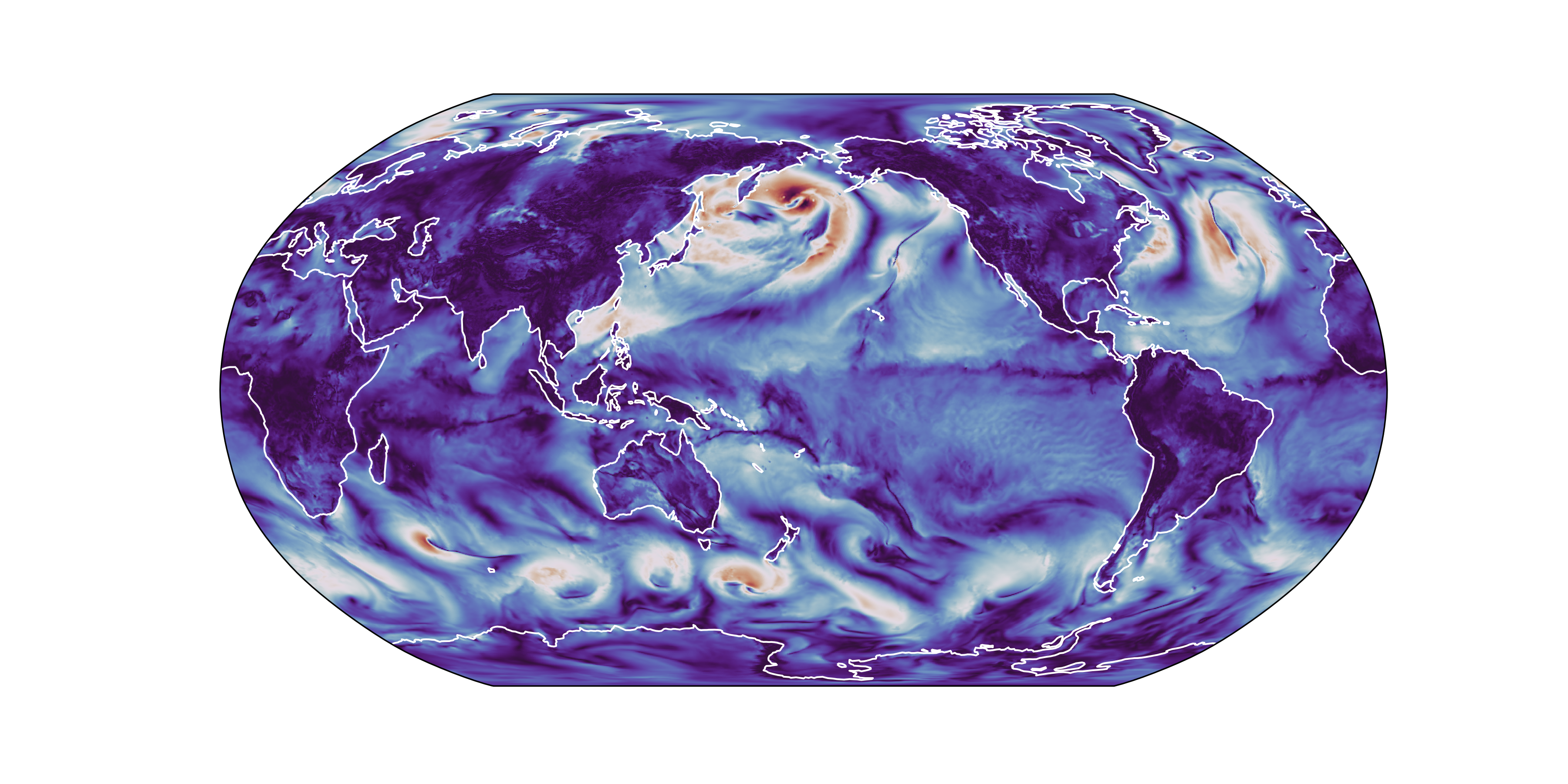}
        \caption{Ground Truth}
        \label{fig:comparison_600h_rollout_gt}
    \end{subfigure}
    \\
    \begin{subfigure}{\linewidth}
        \centering
        \includegraphics[width=0.96\linewidth, trim={100px 40px 100px 40px}, clip]{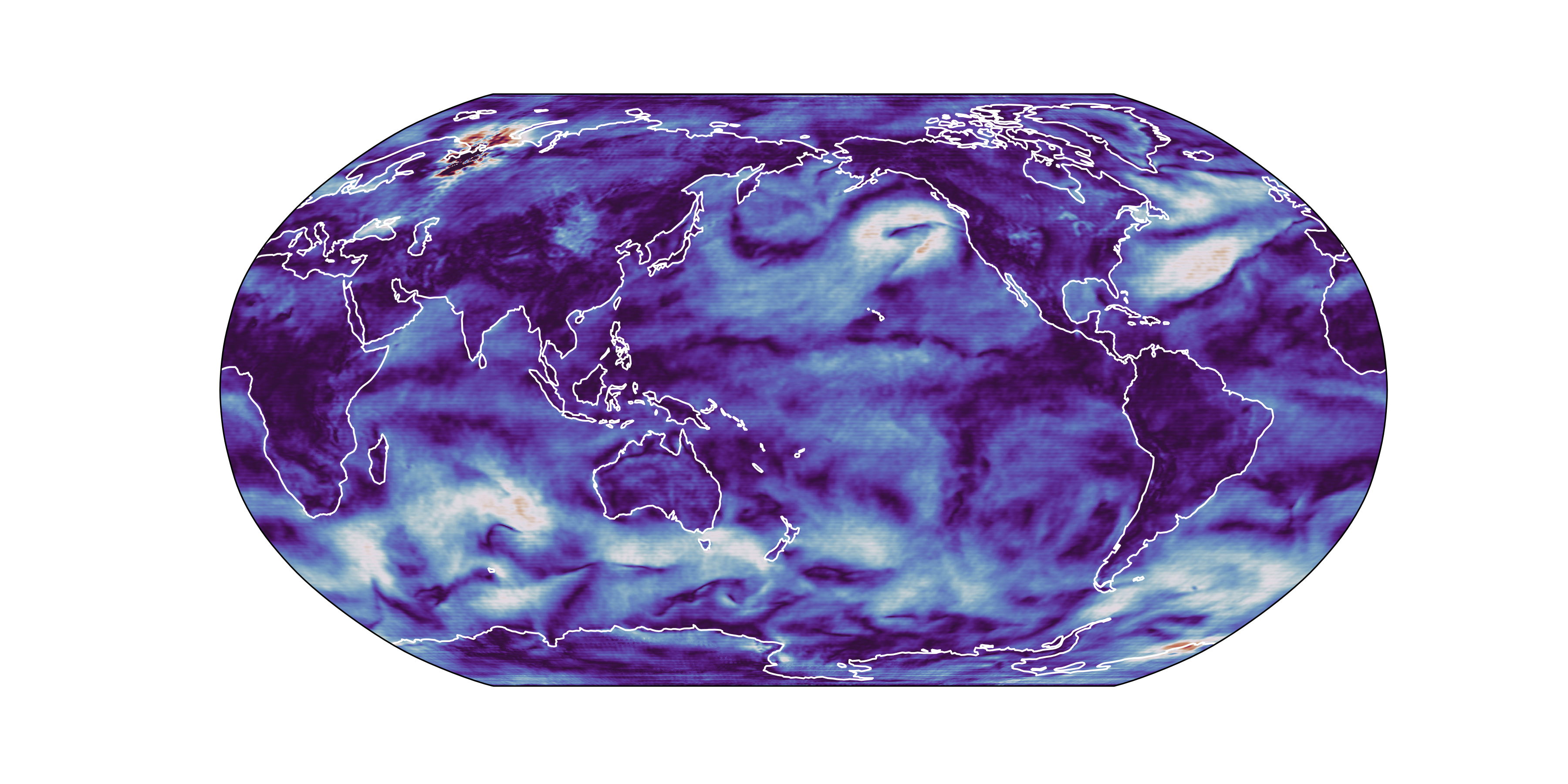}
        \caption{FFT-based FNO architecture for reference}
        \label{fig:comparison_year_rollout_fft}
    \end{subfigure}
    \caption{Year-long rollout (1,450 autoregressive steps) of absolute wind speed 10m above the surface depicting stable behavior over exceptionally long timescales for an ML model, which has important implications for ML-based climate modeling. In contrast, the FFT-based architecture has spurious waves and artifacts and excessive diffusion.
    SFNO faithfully captures the dynamics of weather within the predictability horizon of two weeks and shows physically- and statistically-consistent behaviour over longer timescales up to a year.}
    \label{fig:comparison_year_rollout}
    \vskip -.6cm
\end{figure}

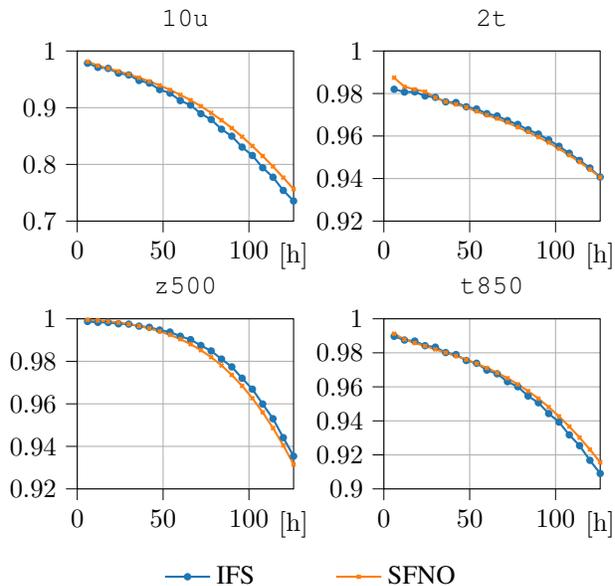
\begin{figure}[!htb]
    \centering
\begin{tikzpicture}

\definecolor{darkgray176}{RGB}{176,176,176}
\definecolor{darkorange25512714}{RGB}{255,127,14}
\definecolor{steelblue31119180}{RGB}{31,119,180}

\begin{groupplot}[group style={group size=2 by 2, vertical sep=1.3cm, horizontal sep=1.2cm}, width=.54\linewidth, grid=both]
\nextgroupplot[
tick align=outside,
tick pos=left,
title={\texttt{10u}},
x grid style={darkgray176},
xmin=0, xmax=126,
xtick style={color=black},
xlabel={$[\mathrm{h}]$},
x label style={at={(1,0.16)}},
y grid style={darkgray176},
ymin=0.70, ymax=1.0,
ytick style={color=black},
]
\addplot [thick, mark=*, mark size=1pt, steelblue31119180]
table[x expr=6+\thisrowno{0}*6, y index=1] {%
0 0.978417217731476
1 0.971107244491577
2 0.969255924224854
3 0.960858762264252
4 0.957716882228851
5 0.947929799556732
6 0.943291008472443
7 0.931923449039459
8 0.925644159317017
9 0.912523210048676
10 0.904474198818207
11 0.889398634433746
12 0.879161298274994
13 0.862030446529388
14 0.849724769592285
15 0.830469787120819
16 0.815665662288666
17 0.794120967388153
18 0.777394533157349
19 0.754106402397156
20 0.735481083393097
};
\addplot [thick, mark=x, mark size=1pt, darkorange25512714]
table[x expr=6+\thisrowno{0}*6, y index=1] {%
0 0.981258034706116
1 0.974306643009186
2 0.969215333461761
3 0.964315414428711
4 0.958722352981567
5 0.952787935733795
6 0.946430742740631
7 0.939348518848419
8 0.931394457817078
9 0.922741711139679
10 0.913235425949097
11 0.902591049671173
12 0.890838027000427
13 0.877905905246735
14 0.864018261432648
15 0.848807692527771
16 0.832515478134155
17 0.814742207527161
18 0.796264171600342
19 0.776727616786957
20 0.756548881530762
};

\nextgroupplot[
tick align=outside,
tick pos=left,
title={\texttt{2t}},
x grid style={darkgray176},
xmin=0, xmax=126,
xtick style={color=black},
x label style={at={(1,0.16)}},
xlabel={$[\mathrm{h}]$},
y grid style={darkgray176},
ymin=0.92, ymax=1,
ytick style={color=black}
]
\addplot [thick, mark=*, mark size=1pt, steelblue31119180]
table[x expr=6+\thisrowno{0}*6, y index=1] {%
0 0.982001066207886
1 0.980690956115723
2 0.980777680873871
3 0.978850603103638
4 0.978203594684601
5 0.976078689098358
6 0.975762367248535
7 0.973763406276703
8 0.972715973854065
9 0.970526695251465
10 0.969447195529938
11 0.967268884181976
12 0.965468049049377
13 0.962932288646698
14 0.960911393165588
15 0.958194971084595
16 0.955201804637909
17 0.951924562454224
18 0.94852751493454
19 0.944989860057831
20 0.940721809864044
};
\addplot [thick, mark=x, mark size=1pt, darkorange25512714]
table[x expr=6+\thisrowno{0}*6, y index=1] {%
0 0.987442433834076
1 0.983169555664062
2 0.981942653656006
3 0.980856418609619
4 0.978104114532471
5 0.976169049739838
6 0.974853992462158
7 0.973518788814545
8 0.971610128879547
9 0.969843447208405
10 0.968180477619171
11 0.966366112232208
12 0.964172124862671
13 0.961889803409576
14 0.959533095359802
15 0.956940054893494
16 0.954015970230103
17 0.950972855091095
18 0.947791278362274
19 0.944328129291534
20 0.940594673156738
};

\nextgroupplot[
tick align=outside,
tick pos=left,
title={\texttt{z500}},
x grid style={darkgray176},
xmin=0, xmax=126,
xtick style={color=black},
xlabel={$[\mathrm{h}]$},
x label style={at={(1,0.16)}},
y grid style={darkgray176},
ymin=0.92, ymax=1,
ytick style={color=black}
]
\addplot [thick, mark=*, mark size=1pt, steelblue31119180]
table[x expr=6+\thisrowno{0}*6, y index=1] {%
0 0.998746514320374
1 0.998312413692474
2 0.998278498649597
3 0.997640132904053
4 0.997495293617249
5 0.996542453765869
6 0.996005952358246
7 0.994718968868256
8 0.993765354156494
9 0.991805911064148
10 0.990205109119415
11 0.987449526786804
12 0.984894931316376
13 0.981065809726715
14 0.97739052772522
15 0.972059011459351
16 0.966906249523163
17 0.959912717342377
18 0.953010022640228
19 0.94411438703537
20 0.935367345809937
};
\addplot [thick, mark=x, mark size=1pt, darkorange25512714]
table[x expr=6+\thisrowno{0}*6, y index=1] {%
0 0.999489188194275
1 0.999178469181061
2 0.998758733272552
3 0.998230814933777
4 0.99749493598938
5 0.996592581272125
6 0.995472609996796
7 0.994110286235809
8 0.992437601089478
9 0.990440309047699
10 0.988060593605042
11 0.985249042510986
12 0.981939792633057
13 0.978066146373749
14 0.973599255084991
15 0.968460738658905
16 0.962629735469818
17 0.956019818782806
18 0.948635339736938
19 0.940469086170197
20 0.931535243988037
};

\nextgroupplot[
tick align=outside,
tick pos=left,
title={\texttt{t850}},
x grid style={darkgray176},
xmin=0, xmax=126,
xtick style={color=black},
x label style={at={(1,0.16)}},
xlabel={$[\mathrm{h}]$},
y grid style={darkgray176},
ymin=0.9, ymax=1,
ytick style={color=black},
legend to name=named,
legend style={legend columns=2, draw=none, /tikz/every even column/.append style={column sep=1.0cm} },
]
\addplot [thick, mark=*, mark size=1pt, steelblue31119180]
table[x expr=6+\thisrowno{0}*6, y index=1] {%
0 0.989705204963684
1 0.987506091594696
2 0.986907124519348
3 0.984182357788086
4 0.983336448669434
5 0.980143308639526
6 0.978972733020782
7 0.975473761558533
8 0.973877906799316
9 0.969830453395844
10 0.967663407325745
11 0.963029444217682
12 0.960069417953491
13 0.95459508895874
14 0.950657963752747
15 0.944291412830353
16 0.939268410205841
17 0.93178004026413
18 0.925422012805939
19 0.916800618171692
20 0.909043371677399
};
\addplot [thick, mark=x, mark size=1pt, darkorange25512714]
table[x expr=6+\thisrowno{0}*6, y index=1] {%
0 0.991229653358459
1 0.987908184528351
2 0.985768675804138
3 0.983983278274536
4 0.982015430927277
5 0.98008143901825
6 0.978129148483276
7 0.976019978523254
8 0.973669528961182
9 0.971082270145416
10 0.968252420425415
11 0.965086698532104
12 0.961502134799957
13 0.957499921321869
14 0.953070819377899
15 0.948111236095428
16 0.942656397819519
17 0.936657249927521
18 0.930179953575134
19 0.923178255558014
20 0.915680348873138
};
\addlegendentry{IFS}
\addlegendentry{SFNO}

\end{groupplot}

\node [below] at (current bounding box.south) {\pgfplotslegendfromname{named}};


\end{tikzpicture}

\vspace{-30pt}
    \caption{Comparison of forecast skill (ACC score) as a function of prediction lead time (in hours) between IFS (blue) and the SFNO (orange) for four physical variables: 10u (surface winds in the latitudinal direction); 2t (temperature 2m above the surface); z500 (geopotential height at 500 hPa); and t850 (the temperature at 850 hPa). All curves are computed by averaging the skill over 730 autoregressive rollouts starting at different initial conditions. SFNO closely matches the forecast skill of IFS in the predicted variables.}
    \label{fig:acc_comparison}
\end{figure}

We demonstrate the utility of the proposed method for the task of medium-range weather forecasting (up to two weeks) and long-timescale rollouts (up to 1 year). We train a handful of models on the ERA5 dataset~\cite{hersbach2020era5} on a subset of atmospheric variables sub-sampled at a temporal frequency of 6 hours and at the native spatial resolution of the ERA5 dataset ($0.25$ degrees lat-long). The atmospheric variables used are listed in appendix \ref{sec:ERA5data}.

We use 40 years of ERA5 (1979-2018): 1979-2015 is used for training, 2016 and 2017 are used for validation, hyper-parameter tuning, and model selection, and 2018 is held out as out-of-sample test set. Models are trained following a protocol similar to that outlined in \citet{Pathak2022}: an initial training stage, using a single autoregressive step and a second, fine-tuning stage, in which two autoregressive steps are used. Scalable model parallelism and gradient checkpointing are used to reduce the large memory footprint encountered during autoregressive training. In both training stages, models are trained for 4 hours on 8 NVIDIA DGX machines. On average, this amounts to 40 epochs in the first stage and 5 epochs in the finetuning stage.

\begin{table*}
\caption{Auto-regressive results obtained on the ERA5 weather data. Relative validation losses (lower is better) are averaged over the finetuning steps at 6 and 12h (1 and 2 autoregressive steps). The Anomaly Correlation Coefficient (ACC, higher is better) is reported at 120h (20 autoregressive steps).}
\label{tab:era5_results}
\begin{center}
\begin{small}
\begin{sc}
\addtolength{\tabcolsep}{-0.5pt}
\begin{tabular}{lcccccccccc}
\toprule
\multirow{2}{*}[-1pt]{Model} & \multicolumn{4}{c}{Parameters} & \multicolumn{2}{c}{Loss} & \multicolumn{4}{c}{ACC at 120h (20 steps)}\\
\cmidrule(lr){2-5} \cmidrule(lr){6-7} \cmidrule(lr){8-11}
 & \scriptsize{Layers} & \scriptsize{Embed. Dim.} & \scriptsize{Param. Count} & \scriptsize{Steps} & \scriptsize{$L^2$} & \scriptsize{$L^1$} & \scriptsize{\texttt{10u}} & \scriptsize{\texttt{2t}} & \scriptsize{\texttt{z500}} & \scriptsize{\texttt{u500}}\\
\midrule
\multirow{2}{*}[-1pt]{FNO, linear}     & \multirow{2}{*}[-1pt]{8} & \multirow{2}{*}[-1pt]{64} & \multirow{2}{*}[-1pt]{$2.190 \cdot 10^9$} & 1 & $0.1172$ & $0.0917$ & $0.4896$ & $0.7829$ & $0.7454$ & $0.5631$\\
                                       &                          &                            &            & 2 & $0.1176$ & $0.0888$ & $0.5140$  & $0.8661$ & $0.8039$ & $0.6223$\\
\cmidrule(lr){5-11}
\multirow{2}{*}[-1pt]{FNO, non-linear} & \multirow{2}{*}[-1pt]{8} & \multirow{2}{*}[-1pt]{384} & \multirow{2}{*}[-1pt]{$4.332 \cdot 10^8$} & 1 & $0.0908$ & $0.0648$ & $0.6284$ & $0.8939$ & $0.8533$ & $0.6912$\\
                                       &                          &                            &            & 2 & $0.0935$ & $0.0669$ & $0.6611$ & $0.9072$ & $0.8677$ & $0.7156$\\
\midrule
\multirow{2}{*}[-1pt]{SFNO, linear}     & \multirow{2}{*}[-1pt]{8} & \multirow{2}{*}[-1pt]{384} & \multirow{2}{*}[-1pt]{$8.296 \cdot 10^8$} & 1 & $0.0930$ & $0.0658$ & $0.6271$ & $0.8965$ & $0.8589$ & $0.6924$\\
                                       &                          &                            &            & 2 & $0.0933$ & $0.0660$ & $0.6471$ & $0.9079$ & $0.8728$ & $0.7141$\\
\cmidrule(lr){5-11}
\multirow{2}{*}[-1pt]{SFNO, non-linear} & \multirow{2}{*}[-1pt]{8} & \multirow{2}{*}[-1pt]{384} & \multirow{2}{*}[-1pt]{$4.332 \cdot 10^8$} & 1 & $0.0942$ & $0.0658$ & $0.6079$ & $0.8943$ & $0.8356$ & $0.6616$\\
                                       &                          &                            &            & 2 & $0.0977$ & $0.0700$ & $0.6489$ & $0.9010$ & $0.8598$ & $0.7012$\\
\bottomrule
\end{tabular}
\addtolength{\tabcolsep}{0.5pt}
\end{sc}
\end{small}
\end{center}
\end{table*}

Table \ref{tab:era5_results} reports model parameters and results for the best checkpoint. For all models, thresholding is employed and frequencies above half the sampling frequency are dropped. The embedding dimension is adapted to facilitate multistep training with 2 autoregressive steps without gradient checkpointing on a single NVIDIA A100. The performance is evaluated using the relative $L^2$ and $L^1$ losses (lower is better), as well as the ACC score (higher is better). For a definition of the performance metrics, we refer the reader to Section \ref{sec:perf_evaluation} in the Appendix.

To measure the predictive skill, we compare autoregressive inference results with a linear SFNO model, with predictions obtained from the state-of-the-art model in NWP -- the Integrated Forecasting System (IFS) \cite{ECMWF2021}. To do so, we compute the ACC scores on a sample of 730 2-week-long forecasts on the out-of-sample data from 2018 shown in Figure~\ref{fig:acc_comparison}.

\textbf{Discussion}
Our SFNO architecture has predictive skill comparable to IFS on weather timescales (up to two weeks, as observed in Figure \ref{fig:acc_comparison}), while showing unprecedented long-term stability for a year-long rollout. More importantly, a one year-long rollout of the SFNO is computed in 12.8 minutes on a single NVIDIA A6000 GPU, compared to one hour (wall-clock time) for a year-long simulation of IFS on 1000 dual-socket CPU nodes \cite{Bauer2020}. With the caveat of differing hardware, this corresponds to a speedup of close to 5,000x. 

While both FFT- and SHT-based approaches achieve similar accuracy for medium-range forecasts (5- to 10-day, Table \ref{tab:era5_results}), the advantage of respecting the underlying spherical geometry becomes evident at longer rollouts as illustrated in Figures \ref{fig:comparison_year_rollout} and \ref{fig:era5_t850_southpole}. Figure \ref{fig:era5_t850_southpole} illustrates the importance of equivariance especially well. Many of our architectural choices were motivated by equivariance and are also present in the non-linear FNO model. While the AFNO model shows artifacts early on, these are not as pronounced in the FNO models and absent in the SFNO model.

\section{Conclusion}
\label{sec:conclusion}
This paper demonstrates how FNOs can be extended via the generalized Fourier transform to learn operators that act on Riemannian manifolds and are equivariant with respect to operations of a symmetry group. We apply this concept to $SO(3)$ and present a novel and fully data-driven network architecture called the SFNO, which facilitates the generation of stable long-range forecasts of Earth's complex atmospheric dynamics of a whole year, in 13 minutes on a single GPU. 

The proposed method pushes the frontier of data-driven deep learning for weather and climate prediction because of the following key properties:

\textbf{Respecting spherical geometry} is essential to ensure that topological boundary conditions are realized correctly. It leads to stable, long-range roll-outs without the respective predictions degenerating to implausible distributions of the respective quantities, such as wind speed. This is an essential property for enabling the creation of ML-based digital twins.

\textbf{Grid-invariance} of the architecture allows the model to be applied on arbitrary grids as long as the SHT and its inverse can be formulated. This allows the method to be quickly fine-tuned on new grids and resolutions, enabling training ML models on diverse weather and climate data that exist on different grids and resolutions.

\textbf{Computational efficiency} is tightly connected to an efficient implementation of the respective generalized Fourier transform. This is the case for spherical topologies via the spherical harmonics transform. As a consequence, the proposed method makes long-term weather forecasting significantly more accessible and is hence enabling its democratization: Predictions for an entire year can be computed within 13 minutes on a single NVIDIA RTX A6000 GPU.

The high accuracy, long-term stability, and immense speedup over classical methods bear promises for the application of Spherical Fourier Neural Operators in the holy grail of prediction: sub-seasonal-to-seasonal forecasting. It is foreseeable that such methods could one day lead to ML-based climate prediction.

\section*{Acknowledgements}

We thank ECMWF for enabling this line of research by providing publicly available datasets \cite{hersbach2020era5}.
Moreover, we extend our gratitude to the anonymous reviewers for taking their time and providing us with valuable comments that improved the quality of the paper. 
Finally, we are grateful to our colleagues Kamyar Azizzadenesheli, Noah Brenowitz, Yair Cohen, Jean Kossaifi, Nikola Kovachki, Thomas M\"uller and Mike Pritchard for fruitful discussions and proof-reading the manuscript.

\bibliography{references}
\bibliographystyle{icml2023}

\newpage
\appendix
\onecolumn

\section{Preliminaries}

\subsection{Rotations on the 2-Sphere}

We discuss some basic notions regarding the two-dimensional sphere $S^2$ and the rotation group $SO(3)$.

To define functions on the sphere, we require coordinates. A familiar choice is the parametrization of points $x \in S^2$ in terms of colatitude $\theta \in [0, \pi]$ and longitude $\varphi \in [0, 2 \pi]$. The unit vector $x$ can then be parametrized as $(\cos(\varphi) \sin(\theta), \sin(\varphi) \sin(\theta), \cos(\theta))^T$.

To discuss symmetries, we require the special orthogonal group in three variables $SO(3)$. These are proper rotations in $\mathbb R^3$, characterized by three-by-three matrices of determinant one, whose inverses are their transpose. Any rotation $R \in SO(3)$ can be written in terms of the Eulerian angles $\varphi \in [0, 2 \pi], \theta \in [0, \pi], \psi \in [0, 2 \pi]$, such that
\begin{equation}
    R = R_z(\varphi) R_y(\theta) R_z(\psi),
\end{equation}
where $R_z$ and $R_y$ are rotations around the z- and y-axes:
\begin{equation}
    R_z(\varphi) = \mqty[\cos \varphi & - \sin \varphi & 0 \\ \sin \varphi & \cos \varphi & 0 \\ 0 & 0 & 1], \quad
    R_y(\varphi) = \mqty[\cos \varphi  & 0 & \sin \varphi\\ 0 & 1 & 0 \\ -\sin \varphi & 0 & \cos \varphi ].
\end{equation}
Unlike translations in the plane, rotations do not commute in general, making $SO(3)$ non-abelian. The rotations $R \in SO(3)$ sweep the entire sphere. We can see this by applying the rotation $R$ to the north pole $n = (0, 0, 1)^T$, which yields
\begin{equation}
    Rn = R_z(\varphi) R_y(\theta) R_z(\psi) n = (\cos(\varphi) \sin(\theta), \sin(\varphi) \sin(\theta), \cos(\theta))^T. 
\end{equation}
We observe that the last rotation angle $\psi$ is dropped, illustrating that $S^2$ can be obtained as the quotient of $SO(3)$ and $SO(2)$.

\subsection{Spherical Harmonics}
\label{sec:spherical_harmonics}

We define the inner product on the 2-Sphere $S^2$
\begin{equation}
    \inner{u}{v}_{L^2(S^2)} = \int_{S^2} \overline{u} \, v \; \dd \Omega = \int_0^{2\pi} \int_0^{\pi} \overline{u(\theta, \varphi)} \, v(\theta, \varphi) \, \sin \theta \;\dd \theta\, \dd \varphi,
\end{equation}
where $\dd \Omega$ is the associated Lebesgue measure on the sphere\footnote{The measure $\dd \Omega$ is invariant under rotations in $SO(3)$. The same is true for the inner product.}. This induces the $L^2$ norm
\begin{equation}
    \label{eq:l2_nrom}
    \norm{u}_{L^2(S^2)} = \sqrt{\inner{u}{u}_{L^2(S^2)}},
\end{equation}
as well as the Hilbert space $L^2(S^2)$ of square integrable functions on $S^2$. We introduce the spherical harmonics defined as
\begin{equation}
    \tag{\ref{eq:spherical_harmonics}}
    Y_l^m(\theta, \varphi)=(-1)^m c_l^m P_l^m(\cos \theta) e^{\mathrm{i} m \varphi}=\widehat{P}_l^m(\cos \theta) e^{\mathrm{i} m \varphi},
    \quad c_l^m:=\sqrt{\frac{2 l+1}{4 \pi} \frac{(l-m) !}{(l+m) !}},
\end{equation}
where $P_l^m(\cos \theta)$ are the associated Legendre polynomials. The normalization factor $c^m_l$ normalizes the spherical harmonics w.r.t. the $L^2(S^2)$ inner product, s.t.
\begin{equation}
    \inners2{Y_l^m }{Y_{l'}^{m'} } = \delta_{ll'} \delta_{mm'}.
\end{equation}
In other words, the spherical harmonics \eqref{eq:spherical_harmonics} form an orthogonal basis of $L^2(S^2)$.

The spherical harmonics \eqref{eq:spherical_harmonics} have many useful properties, induced by properties of the trigonometric functions and the associated Legendre polynomials \cite{abramowitz1964handbook}. One such useful property is the symmetry relation
\begin{equation}\label{eq:ylm-symmetry}
\overline{Y_l^m}(\theta, \phi) = (-1)^m \,Y_l^{-m}(\theta, \phi),
\end{equation}
which is particularly useful for real-to-complex spherical harmonic transforms. Using this property, we can recover the negative $m$ components from the positive ones\footnote{Analogous to the Hermitian symmetry of the Fourier coefficients of a real-valued signal where negative frequency contributions can be inferred from the positive ones.}.

\subsection{Fourier Neural Operators}

On the doubly periodic domain $S^1 \times S^1$, the Fourier transform and it's inverse can be expressed as
\begin{equation}
    \mathcal{F}[u](k,l) = \tilde{u}(k,l) = \int_0^1 \int_0^1 u(x, y) \exp(-i 2\pi k x)\dd x \exp(-i 2\pi l y)\dd y
\end{equation}
and 
\begin{equation}
    \mathcal{F}^{-1}[\tilde{u}](x) = \sum_{k=-\infty}^{\infty} \sum_{l=-\infty}^{\infty} \tilde{u}(k,l) \exp(i 2\pi k x)\exp(i 2\pi l y)
\end{equation}
respectively.

At the core of the Fourier Neural Operator lies the Fourier layer, which can be understood as a global convolution
\begin{equation}\tag{\ref{eq:convolution}}
u \rightarrow \mathcal{K}[u] = \mathcal{F}^{-1}[\mathcal{F}[\kappa] \cdot \mathcal{F}[u]].
\end{equation}
By replacing the filter weights $\mathcal{F}[\kappa]$ with learned weights $\tilde{\kappa}_\vartheta$ in Fourier space, we obtain the Fourier Neural Operator
\begin{equation}\label{eq:fno}
u \rightarrow \mathcal{K}_\vartheta[u] = \mathcal{F}^{-1}[\tilde{\kappa}_\vartheta \cdot \mathcal{F}[u]].
\end{equation}
Here, $\tilde{\kappa}_\vartheta$ represents a parametrization of the filter weights in Fourier space with the weight vector $\vartheta$. As such, a filter weight $\tilde{\kappa}_\vartheta(k,l)$ is learned for each frequency $k, l$, such that
\begin{equation}
    \mathcal{K}_\vartheta[u] = \mathcal{F}^{-1}[\tilde{\kappa}_\vartheta \cdot \tilde{u}](x) = \sum_{k=-\infty}^{\infty} \sum_{l=-\infty}^{\infty} \tilde{\kappa}_\vartheta(k,l) \cdot \mathcal{F}[u](k,l) \; \exp(i 2\pi k x)\exp(i 2\pi l y).
\end{equation}
In practice, we truncate the Fourier series as only a finite number of filter parameters is learned. Moreover, if $u$ is vector-valued, we replace $\tilde{\kappa}_\vartheta(k,l)$ with a learned matrix and $\tilde{\kappa}_\vartheta(k,l) \cdot \mathcal{F}[u]$ becomes a matrix-vector product with summation over the embedding dimension \cite{FNO}.

The above formulation admits efficient learning of non-local operators, which is a desireable property for PDE applications \cite{Kovachki2021}. We remark that the FNO is trivially translation-equivariant w.r.t. translations $T(a,b): (x, y) \rightarrow (x+a, y+b)$. Applying the passive translation $\Phi_T[u](x) = u(T^{-1}x)$ to the input of the FNO yields
\begin{align}
    \mathcal{K}_\vartheta[\Phi_T[u]] &= \mathcal{F}^{-1}[\tilde{\kappa}_\vartheta \cdot \mathcal{F}[\Phi_T[u]]] = \mathcal{F}^{-1}[\tilde{\kappa}_\vartheta \cdot \exp(i 2\pi k a) \exp(i 2\pi l b)\,\mathcal{F}[u]] \nonumber \\
    &= \Phi_T[\mathcal{F}^{-1}[\tilde{\kappa}_\vartheta \cdot \mathcal{F}[u]]] = \Phi_T[\mathcal{K}_\vartheta[u]],
\end{align}
where we have used $\Phi_T[\mathcal{F}[u]] = \exp(i 2\pi k a) \exp(i 2\pi l b)\,\mathcal{F}[u]$.

\section{Implementation Details}

\subsection{Differentiable Spherical Harmonics Transforms}
\label{sec:differentiable_sht}
In order to facilitate differentiable computation we require a differentiable implementation of the Spherical Harmonic Transforms (SHT). To do so we implemented our own
differentiable Spherical Harmonics Transform library for \texttt{PyTorch} (\url{https://pytorch.org/}).

While it is possible to compute the SHT in $\mathcal{O}(L^2 \log^2 L)$ time \cite{Driscoll1994, Schaeffer2013, McEwen2011}, these algorithms typically suffer from numerical-intabilities and tend to have large constants in their algorithmic complexity. As such, the 'semi-naive' algorithm, which computes the projection onto the associated Legendre polynomials via quadrature and the projection onto the harmonic functions via the FFT \cite{Schaeffer2013}. In our experience, this algorithm tends to outperform the asymptotically optimal algorithms, and moreover, is better-suited to the execution on GPUs and distributed GPU-systems due to the availability of highly optimized primitives for these operations \cite{cuda}. As such, 
\texttt{torch-harmonics} implements the 'semi-naive' algorithms and distributes the projection of the associated Legendre polynomials onto multiple GPUs in the distributed case.

We implement the forward transformation in equation (\ref{eq:sh-transform}) by performing a 1D real to complex DFT over the azimuth degrees of freedom, followed by a matrix multiplication of the Legendre polynomials $P_l^m(\cos\theta)$ and re-scaled by the normalization factors $c_{l}^m$. Therefore, we can write
\begin{equation}\label{eq:dsht-forward}
\hat{u}[\ldots, l, m] = \sum\limits_{j=0}^{H-1} P[l, m, j]\,\mathrm{FT}\big[u[\ldots, j, k], k\big][j, m]
\end{equation}
Where $0\leq j< H$, $0\leq k < W$, where $H$ and $W$ are the number of discrete $\theta$ and $\phi$ angles respectively.
Furthermore, $0\leq l \leq L$ and $0\leq m \leq M$. Note that we do not need to store the modes for negative $m$ since we are only considering real-to-complex transforms and thus those components can be retrieved using the symmetry relation (\ref{eq:ylm-symmetry}).
Lastly, the ellipsis in (\ref{eq:dsht-forward}) denotes all tensor modes which are not contracted in the transform and thus can be vectorized over. Typical modes include the batch as well as the feature dimension of tensor u. 

We define the real-to-complex DFT as
\begin{equation}
\mathrm{FT}\big[u[\ldots, j, k], k\big][j,m] \equiv \frac{1}{\sqrt{W}}\sum\limits_{k=0}^{W-1} \exp(-i\frac{2\pi m k}{W}) u[\ldots, j, k]
\end{equation}
Note that the negative phase arises from the fact that the forward transform involves the complex conjugate of $Y_l^m$.
We define the Legendre weights as
\begin{equation}
P[l, m, j] \equiv (-1)^m\,c_l^m\,P_l^m(\cos\theta_j)\,w(\theta_j),
\end{equation}
where the weight vector
\begin{equation}
w(\theta_j) \simeq \sin\theta \cdot \mathrm{d}\theta,
\end{equation}
is chosen such that the sum over $j$ in (\ref{eq:ylm-symmetry}) approximates the integral over $\theta$ in (\ref{eq:sh-transform}). These weight factors typically depend on the chosen grid of the spatial discretization.
Since we are working with static input and output grids, the Legendre weight matrices $P$ can be pre-computed and stored.
Furthermore, effective downsampling can be achieved by reducing the maximum wave number $L$. 

The inverse transform can be defined in a similar fashion:
\begin{equation}\label{eq:dsht-backward}
u[\ldots, j, k] = \mathrm{FT}^{-1}\left[\sum\limits_{l=0}^{L} \hat{P}[l, m, j]\,\hat{u}[\ldots, l, m], m \right][j, k]
\end{equation}
where $\mathrm{FT}^{-1}\big[\ldots, m\big][j, k]$
denotes the inverse complex-to-real inverse DFT with respect to $m$.
The Legendre weights in this case are given by:
\begin{equation}
\hat{P}[l, m, j] \equiv (-1)^m\,c_l^m\,P_l^m(\cos\theta_j).
\end{equation}
Note that we do not need to include an additional weight vector for the backward transform since the sum over $l$ in (\ref{eq:dsht-backward}) does not approximate an integral and thus does not receive an additional term from an integration measure. 


The above implementation of discrete SHT can be parallelized using the pencil decomposition technique which is also used in higher-dimensional distributed Fourier transforms: first, the user needs to initializes a GPU communication grid for $H,L$ and $W,M$ directions respectively and pass this to the distributed initialization routine of \texttt{torch-harmonics}. 
We further assume that the input to the distributed transform is split evenly among those communication dimensions (including padding if required).

For example, in case of the forward transform, the input to the distributed transform can be a spatially decomposed field in $H$ and $W$ but with local feature data. The idea is to perform a global transposition using \texttt{torch.distributed.all\_to\_all} of the data in feature and $W$ domain, so that after the transposition the $W$ domain is fully local and the feature domain is distributed. After this, the one dimensional FFT along $W$ can be performed locally and is embarrassingly parallel with respect to the now split feature dimension. After this first transform, the data is again globally transposed such that the $H$ domain local and the the feature domain as well as the $m$ domain are both distributed. In This case, the Legendre transformation can be performed locally and embarrassingly parallel with respect to the distributed feature dimension. Afterwards, we perform a third global transposition to achieve a decomposed $L,M$ domain but local feature data. For the inverse SHT, we invert this whole process. Output dimensions which cannot be evenly split among GPUs, are zero-padded automatically on the largest rank in the corresponding communication dimension.
Information about the padding and input and output sizes are stored in the corresponding SHT instance and can be queried by the user.
The advantage of this approach is that input and output tensors have similar spatial decompositions and thus this approach simplifies the end-to-end spatial parallelization of SFNO models.
Additionally, since all transforms are performed locally, the result of the distributed spherical transform is bit-wise identical to the result of a serial transform. This is not the case for fork-join approaches as discussed below.

Alternatively, one could also follow a fork-join approach
in forward and inverse SHT respectively. This is can be achieved by splitting the l degrees-of-freedom of the Legendre weights $P, \hat{P}$ into evenly sized chunks $l_c \doteq (L{+}1) / R$ and distributing those across all ranks $R$. This means, that every rank $0\leq r < R$ owns a sub-tensor $P_r[l, m, j]$ with $r\cdot l_c \leq l < (r+1)\cdot l_c$ but full $m$ and $j$ (and equivalently for $\hat{P}$). 
Therefore, the distributed forward transform variant of (\ref{eq:dsht-forward}) transforms a shared input tensor $u[\ldots, j, k]$ onto a rank-local transformed tensor $u_r[\ldots, l, m]$ with the above ranges for $l, m$ (fork). The distributed inverse transform in turn transforms the rank-local tensor $\hat{u}$ back into a shared tensor $u$ (join). 
The communication primitive for both operations is \texttt{torch.distributed.all\_reduce}, where it has to be applied on the input gradient in the backward pass of the forward routine as well as on the output tensor in the forward pass of the inverse routine. The forward pass of the forward and the backward pass of the inverse transformation do not need additional communication.
Since PyTorch cannot back-propagate through communication collectives natively, we use \texttt{torch.autograd.Function} to implement collective primitives with fully defined forward and backward pass.

\subsection{Training}
\label{sec:training}


To find the parameter vector $\vartheta$ of the learned map $F_\vartheta$, an objective function is minimized. We choose a geometric loss function, which is obtained by approximating the $L^p$ norm $\norm{F_\vartheta[\vb{u}_n] - \vb{u}_{n+1}}_{L^p(S^2)}$ on the sphere:
\begin{equation}
    \label{eq:lp_loss}
    \mathcal{L}
    [F_\vartheta[\vb{u}_n], \vb{u}_{n+1}]
    = \sum_{c \in \text{channels}} \left(\frac{\sum_{i \in \text{grid}} w_i \left| F_\vartheta[\vb{u}_n](\vb{x}_i) - \vb{u}_{n+1}(\vb{x}_i) \right|^p}{\sum_{i \in \text{grid}} w_i \left|\vb{u}_{n+1}(\vb{x}_i)\right|^p} \right)^{\frac{1}{p}}.
\end{equation}
To compute the loss, the absolute difference is summed over the gridpoints $x_i$ and weighted with $w_i$, which are the products of the Jacobian $\sin \theta_i$ and the quadrature weights. As quadrature rule we picked the simple Riemann sum\footnote{On equiangular grids on the sphere, Clenshaw-Curtiss quadrature is often preferable \cite{Schaeffer2013}. This did not improve performance over the Riemann sum however, and we chose to use the simpler Riemann sum instead.}
The loss \eqref{eq:lp_loss} is then computed for each channel $c \in \{\texttt{u10m}, \texttt{v10m}, \texttt{t2m}, \ldots\}$ separately, and then normalized by the norm of the target. It is then averaged over all predicted channels to obtain the final loss. For training we set $p=2$.

Training is performed in two stages. In the first stage, the model is trained to obtain the best possible single-step performance. To this end, the loss \eqref{eq:lp_loss} is minimized after a single prediction step. The learning rate is scheduled to follow a cosine pattern, starting with a learning rate of $10^{-3}$

The second stage is a finetuning stage where the model is optimized for autoregressive performance. To this end, $n_\text{steps}$ autoregressive steps are performed and the loss is accumulated at each step of the forecast, i.e.
\begin{equation}
    \label{eq:autoregressive_loss}
    \mathcal{L}_\text{autoregressive} = \frac{1}{n_\text{steps}} \sum_{s=1}^{n_\text{steps}}\mathcal{L}[F^s_\vartheta[\vb{u}_n], \vb{u}_{n+m}], \quad \text{where} \quad F^s_\vartheta = \underbrace{F_\vartheta \circ \ldots \circ F_\vartheta}_\text{$s$ times}.
\end{equation}
Gradients are then backpropagated through the entire unrolled sequence, to get the weight updates. The finetuning is performed with with increasing $n_\text{steps}$, starting at $n_\text{steps}=2$. For each $n_\text{steps}$, the model is trained for 5 epochs with a constant learning rate of $10^{-5}$.

\subsection{Performance evaluation}
\label{sec:perf_evaluation}

To assess the performance of our models, the we use the relative $L^2$ and $L^1$ losses as defined in equation \eqref{eq:lp_loss}. Another common metric used in weather prediction is the anomaly correlation coefficient (ACC). The latitude weighted ACC for a forecast variable $v$ at forecast time-step $l$ is defined following~\citet{rasp2020weatherbench} as follows:

\begin{align}
\label{eq:acc}
    \mathrm{ACC}(v, l)=\frac{\sum_{m, n} w_\text{Lat}(m) \tilde{{X}}_{\text{pred}}(l)\left[ v, m, n \right] \tilde{{X}}_{\text{true}}(l)\left[v, m, n\right]}{\sqrt{\sum_{m, n} w_\text{Lat}(m) \left( \tilde{{X} }_{\text{pred}}(l)\left[v, m, n\right]\right)^{2} \sum_{m, n} w_\text{Lat}(m) \left(\tilde{{X}}_{\text{true}}(l)\left[v, m, n\right]\right)^{ 2}}},
\end{align}

where $\tilde{{X}}_{\text{pred/true}}(l)\left[v, m, n \right]$ represents the long-term-mean-subtracted value of predicted (/true) variable $v$ at the location denoted by the grid co-ordinates $(m, n)$ at the forecast time-step $l$. The long-term mean of a variable is simply the mean value of that variable over a large number of historical samples in the training dataset. The long-term mean-subtracted variables $\tilde{{X}}_{\text{pred/true}}$ represent the anomalies of those variables that are not captured by the long term mean values. $w_\text{Lat}(m)$ is the latitude weighting factor at the co-ordinate $m$. The latitude weighting is defined by Equation~\ref{eq:latweight} as
\begin{align}
\label{eq:latweight}
w_\text{Lat}(j)=\frac{\cos (\operatorname{Lat}(m))}{\frac{1}{N_{\text {Lat }}} \sum_{j}^{N_{\text {lat }}} \cos (\operatorname{Lat}(m))}.
\end{align} We report the mean ACC over all computed forecasts from different initial conditions and report the variability in the ACC over the different initial conditions by showing the first and third quartile value of the ACC in all the ACC plots that follow unless stated otherwise.

\section{Datasets}
\label{sec:datasets}

\subsection{Shallow Water Equations on the Rotating Sphere}
\label{app:swe_experiments}

The shallow water equations on the rotating 2-sphere model a thin layer of fluid covering a rotating sphere. They are typically derived from the three-dimensional Navier-Stokes equations, assuming incompressibility and integrating over the depth of the fluid layer. They are formulated as a system of hyperbolic partial differential equations
\begin{equation}
    \begin{cases}
        \partial_t \varphi +\nabla \cdot(\varphi\vb{u}) = 0 & \text { in } S^2 \times(0, \infty) \\
        \partial_t (\varphi\vb{u}) +\nabla \cdot \vb{T} =\vb{S} & \text { in } S^2 \times(0, \infty) \\
        \varphi=\varphi_0 & \text { on } S^2 \times\{t=0\}, \\
        \vb{u}=\vb{u}_0 & \text { on } S^2 \times\{t=0\}.
    \end{cases}
    \label{eq:swe}
\end{equation}
The state vector $(\varphi, \varphi u^T)^T$ contains the geopotential layer depth $\varphi$ (mass) and the tangential momentum vector $\varphi u$ (discharge). In curvilinear coordinates, the flux tensor $T$ can be written written with the outer product as $\varphi \vb{u} \otimes \vb{u}$. The right-hand side contains flux terms such as the Coriolis force. A detailed treatment of the SWE equations can be found in e.g. \cite{Giraldo2001, Bonev2018, Nair2005}.

Training data for the SWE is generated by randomly generating initial conditions and advancing them in time using a classical numerical solver. The initial geopotential height and velocity fields are realized as Gaussian random fields on the sphere. The initial layer depth has an average of $\varphi_\text{avg} = 10^3 \cdot g$ with a standard deviation of $120 \cdot g$. The initial velocity components have a zero mean and a standard deviation of $0.2 * \sqrt{\varphi_\text{avg}}$. The parameters of the PDE, such as gravity, radius of the sphere and angular velocity, we choose the parameters of the Earth. Training data is generated on the fly by using a spectral method to numerically solve the PDE on an equiangular grid with a spatial resolution of $256 \times 512$ and timesteps of 150 seconds. Time-stapping is performed using the third-order Adams-Bashford scheme. The numerical method then computes geopotential height, vorticity and divergence as output. 

This data is z-score normalized and the modes are trained using epochs containing 256 samples each. To optimize the weights, we use the popular Adam optimizer with a learning rate of $2\cdot10^{-3}$.

\subsection{Weather prediction/ERA5 Data}
\label{sec:ERA5data}

It is a multi-decadal, high-frequency estimate of the state of the Earth’s atmosphere. It is the result of reanalysis, a process that uses data-assimilation~\cite{carrassi2018data, kalnay2003atmospheric} to combine modern numerical weather forecasting models with historical observational records to produce an estimate of the historical ocean-atmosphere system. A reanalysis dataset such as ERA5 spans multiple decades using dynamics from an unchanging modern numerical model. This is in contrast to an operational analysis dataset where the numerical model gets periodically updated due to advances in computational techniques, numerical methods and and improvements in the understanding of fundamental  geophysics. Thus a reanalysis dataset maintains temporal consistency. Furthermore, raw observations of the earth’s ocean-atmosphere system are sparsely distributed in time and space, multimodal, and of variable quality. A reanalysis dataset assimilates various observational sources informed by the uncertainty estimates of those observation sources. A reanalysis dataset provides a consistent picture of the history of the earth’s atmosphere making it very useful as a training dataset for a machine learning model. Consequently, a large and growing number of researchers~\cite{rasp2021data, weyn2020improving, scher2019weather, arcomano2020machine, Pathak2022, Bi2022, Lam2022} have used the ERA5 dataset for training data-driven numerical weather models. 


Models are trained on two subsets of the variables: a 26 channel dataset for evaluating and comparing models to each other, and a 73 channel dataset used to train a larger model for comparison with IFS. The following set of 26 variables is used in the 26 variable dataset: \texttt{10u}, \texttt{10v}, \texttt{2t}, \texttt{sp}, \texttt{msl}, \texttt{tcwv}, \texttt{100u}, \texttt{100v},
\texttt{z50}, \texttt{z250}, \texttt{z500}, \texttt{z850}, \texttt{z1000}, \texttt{u250}, \texttt{u500}, \texttt{u850}, \texttt{u1000}, \texttt{v250}, \texttt{v500}, \texttt{v850}, \texttt{v1000},
\texttt{t100}, \texttt{t250}, \texttt{t500}, \texttt{t850},
\texttt{r500}.
For the 73 variable dataset, we add \texttt{z---}, \texttt{t---}, \texttt{u---}, \texttt{v---} and
\texttt{r---} at pressure levels 50, 100, 150, 200, 250, 300, 400, 500, 600, 700, 850, 925, 1000 hPa to the already existing variables. 
Table \ref{tab:era5_variables} lists an overview of variables used during training and their meaning. For a complete overview, we refer to the ECMWF website \url{https://apps.ecmwf.int/codes/grib/param-db}.

\begin{table}[t]
\caption{Atmospheric and Surface variables predicted by our model. Two subsets of 26 and 73 variables each are used to train models of various sizes. Detailed desciptions can be accessed on the ECMWF website \url{https://apps.ecmwf.int/codes/grib/param-db}.}
\label{tab:era5_variables}
\vskip 0.15in
\begin{center}
\begin{small}
\begin{sc}
\begin{tabular}{llccc}
\toprule
Desn  & Description & ECMWF ID \\
\midrule
\texttt{10u}    & 10 metre $u$-wind component                                   & 165       \\
\texttt{10v}    & 10 metre $v$-wind component                                   & 166       \\
\texttt{2t}     & 2 metre temperature                                           & 167       \\
\texttt{sp}     & Surface pressure                                              & 135       \\
\texttt{msl}    & Mean sea level pressure                                       & 151       \\
\texttt{tcwv}   & Total column vertically-integrated water vapour               & 137       \\
\texttt{100u}   & 100 metre $u$-wind component                                  & 228246    \\
\texttt{100v}   & 100 metre $v$-wind component                                  & 228247    \\
\texttt{z---}   & Geopotential  (at pressure level \texttt{---})                & 129       \\
\texttt{t---}   & Temperature (at pressure level \texttt{---})                  & 130       \\
\texttt{u---}   & $u$ component of the wind (at pressure level \texttt{---})    & 131       \\
\texttt{v---}   & $v$ component of the wind (at pressure level \texttt{---})    & 132       \\
\texttt{r---}   & Relative humidity (at pressure level \texttt{---})            & 157       \\
\bottomrule
\end{tabular}
\end{sc}
\end{small}
\end{center}
\vskip -0.1in
\end{table}

\section{Supplementary material}
\label{sec:supplementary_material}
Our differentiable implementation of the SHT and a reference implementation of SFNO can be found at \url{https://github.com/NVIDIA/torch-harmonics}.

Short videos depicting long rollouts of SFNO and polar artifacts can be found at
\url{https://youtu.be/OM3JZZN5uE4} and \url{https://youtu.be/LPVejeU8YDE}, respectively.


\end{document}